\documentclass{article}

\usepackage{microtype}
\usepackage{graphicx}
\usepackage{subcaption}
\usepackage{booktabs} 

\usepackage{hyperref}

 \usepackage[preprint]{icml2026}

\usepackage{listings}
\usepackage{amsmath}
\usepackage{amssymb}
\usepackage{mathtools}
\usepackage{amsthm}
\usepackage{comment}
\usepackage{algorithm}
\usepackage{algorithmic}
\usepackage{tikz}
\usetikzlibrary{arrows.meta}
\usetikzlibrary{shadows.blur}
\usetikzlibrary {decorations.pathmorphing}
\usetikzlibrary{decorations.pathreplacing}
\usetikzlibrary{shapes}
\usepackage{multirow}
\usepackage{graphicx}
\usepackage{adjustbox} 
\usetikzlibrary{
  positioning,fit,arrows.meta,calc,
  shadows.blur,backgrounds,
  decorations.pathreplacing,
  shapes.geometric
}
\usepackage[capitalize,noabbrev]{cleveref}
\usepackage{amsmath,amssymb}

\newcommand{\clip}{\mathrm{clip}}

\newcommand{\SOL}{\mathrm{SOL}}
\theoremstyle{plain}

\theoremstyle{definition}

\theoremstyle{remark}

\usepackage{xcolor}

\definecolor{rankA}{HTML}{32CD32} 
\definecolor{rankB}{HTML}{F0E130} 
\definecolor{rankC}{HTML}{FDBCB4} 

\newcommand{\best}[1]{\colorbox{rankA}{#1}} 
\newcommand{\second}[1]{\colorbox{rankB}{#1}} 
\newcommand{\third}[1]{\colorbox{rankC}{#1}} 

\usepackage[textsize=tiny]{todonotes}

\begin{document}

\twocolumn[
  \icmltitle{OptiML: An End-to-End Framework for Program Synthesis and CUDA Kernel Optimization}



  \icmlsetsymbol{equal}{*}

  \begin{icmlauthorlist}
    \icmlauthor{Arijit Bhattacharjee}{iowa}
    \icmlauthor{Heng Ping}{sch}
    \icmlauthor{Son Vu Le}{iowa}
    \icmlauthor{Paul Bogdan}{sch}
    \icmlauthor{Nesreen K. Ahmed}{comp}
    \icmlauthor{Ali Jannesari}{iowa}
  \end{icmlauthorlist}

  \icmlaffiliation{iowa}{Iowa State University, Ames, Iowa, USA}
  \icmlaffiliation{comp}{Cisco AI Research, San Jose, California, USA}
  \icmlaffiliation{sch}{University of Southern California, Los Angeles, California, USA}

  \icmlcorrespondingauthor{Arijit Bhattacharjee}{arbhatt9@iastate.edu}

  \icmlkeywords{Machine Learning, ICML}

  \vskip 0.3in
]



\printAffiliationsAndNotice{}  

\begin{abstract}
Generating high-performance CUDA kernels remains challenging due to the need to navigate a combinatorial space of low-level transformations under noisy and expensive hardware feedback. Although large language models can synthesize functionally correct CUDA code, achieving competitive performance requires systematic exploration and verification of optimization choices. We present OptiML, an end-to-end framework that maps either natural-language intent or input CUDA code to performance-optimized CUDA kernels by formulating kernel optimization as \emph{search under verification}.

\textsc{OptiML} consists of two decoupled stages. When the input is natural language, a Mixture-of-Thoughts generator (\textsc{OptiML-G}) acts as a proposal policy over kernel implementation strategies, producing an initial executable program. A search-based optimizer (\textsc{OptiML-X}) then refines either synthesized or user-provided kernels using Monte Carlo Tree Search over LLM-driven edits, guided by a hardware-aware reward derived from profiler feedback. Each candidate transformation is compiled, verified, and profiled with Nsight Compute, and evaluated by a composite objective that combines runtime with hardware bottleneck proxies and guardrails against regressions. We evaluate \textsc{OptiML} in both synthesis-and-optimize and optimization-only settings on a diverse suite of CUDA kernels. Results show that \textsc{OptiML} consistently discovers verified performance improvements over strong LLM baselines and produces interpretable optimization trajectories grounded in profiler evidence.
\end{abstract}

\section{Introduction}

Modern accelerator programming remains a central bottleneck for scientific computing and machine learning systems~\cite{chen2018tvm,zheng2020ansor}. While GPUs offer immense throughput, achieving high performance still demands expertise in memory hierarchies, execution configuration, synchronization, and compiler behavior~\cite{dao2022flashattention}. The gap between \emph{functionally correct} kernels and \emph{performance-portable} kernels is particularly wide: small code decisions (e.g., access patterns, reuse strategy, and instruction mix) can shift a kernel from compute-bound to memory-bound or vice versa, and the appropriate optimization depends on both hardware and workload characteristics~\cite{ouyang2025kernelbenchllmswriteefficient, rodriguez2025computeeval}.
This challenge is amplified in emerging workflows where kernels are produced (or edited) by large language models (LLMs)~\cite{chen2021evaluatinglargelanguagemodels,jiang2024survey}. Although LLMs can often synthesize syntactically valid CUDA code, the resulting kernels may fail to compile, violate numerical constraints, or leave substantial performance untapped due to missed low-level transformations~\cite{chen2025cudallmllmswriteefficient, nichols2024large, ouyang2025kernelbenchllmswriteefficient}. More broadly, existing research treats code generation and optimization as separate problems: prior work either translates high-level programs (e.g., C++ or Python) into CUDA~\cite{tehrani2024coderosetta,ke2025qimengmupamutualsupervisedlearningsequentialtoparallel,pmlr-v162-wen22b} or tunes existing kernels through autotuning~\cite{HOZZOVA2021107631,bjertnes2021lscatlargescalecudaautotuning} and heuristics. In contrast, we study an end-to-end setting in which the input may be either natural-language intent or user-provided code and the output is a performance-optimized CUDA kernel.

A natural response is to apply automated optimization after code generation~\cite{ye2025promptalchemyautomaticprompt}. However, existing post-hoc optimization approaches face two fundamental difficulties. First, the optimization space for GPU kernels is combinatorial and non-smooth: changes such as tiling, vectorization, shared-memory staging, or loop restructuring interact in complex ways with register pressure, occupancy, and memory coalescing. Second, performance feedback is expensive and noisy; relying solely on end-to-end timing can obscure the root cause of inefficiency and makes search brittle~\cite{pragma2025, tschand2025swizzleperfhardwareawarellmsgpu}. As a result, naive strategies, such as random perturbations, greedy edits, or fixed rule-based passes, often converge slowly, overfit to narrow kernel classes, or fail to generalize across diverse patterns~\cite{ye2025promptalchemyautomaticprompt, guo2025evoengineermasteringautomatedcuda}.

In this paper, we present \textbf{OptiML}, an end-to-end framework that unifies program synthesis and kernel optimization, directly addressing the separation between generation and tuning in prior work. Unlike existing approaches that either translate high-level programs into CUDA or optimize pre-existing kernels in isolation, OptiML maps either natural-language intent or input CUDA code to performance-optimized kernels by formulating kernel optimization as a \emph{guided decision process grounded in hardware behavior}~\cite{li2025autotritonautomatictritonprogramming, baronio2025kevinmultiturnrlgenerating}. OptiML consists of two decoupled components. \textbf{OptiML-G} is a Mixture-of-Thoughts generator~\cite{feinashley2025mot} that acts as a \emph{proposal policy generator} over kernel implementation strategies when the input is natural language, synthesizing an initial executable program. \textbf{OptiML-X} is a search-based optimization engine that refines either synthesized or user-provided kernels by exploring program transformations using \emph{Monte Carlo Tree Search (MCTS)}~\cite{uct,lim2016mcts} and an \emph{LLM-as-a-Judge}~\cite{zheng2023judging} to propose and evaluate candidate edits at test time~\cite{wang2025mctsjudge, mctsahd2025}. Rather than treating optimization as black-box tuning, OptiML-X leverages profiler-derived proxy metrics (e.g., utilization and memory-traffic surrogates) to identify performance bottlenecks~\cite{bhattacharjee2023openmp} and prioritize transformations that target them. This framing makes the search both more sample-efficient and more interpretable: the system can attribute improvements to reduced memory traffic, improved utilization, or lower instruction footprint, and can avoid edits that trade one bottleneck for another.

Post-hoc optimization alone may be helpful but often insufficient when the initial program structure is poor, incomplete, or fails to compile, as search-based refinement cannot reliably recover high-performance kernels from weak starting points. OptiML addresses this limitation by augmenting optimization with a proposal generator that generates optimization-ready and structurally strong kernels from natural language intent. In this role, \textbf{OptiML-G} provides a structured initialization by routing among latent implementation strategies to synthesize an executable kernel that exposes meaningful opportunities for transformation.

The full end-to-end proposed framework, \textbf{OptiML (OptiML-G + OptiML-X)}, therefore unifies program synthesis and optimization into a single end-to-end pipeline. When the input is natural language, OptiML-G supplies a strong starting point that enables effective downstream search. When the input is user-provided code, OptiML bypasses generation and directly applies the same profiler-grounded optimization procedure. In both cases, performance improvements are driven by verified transformations grounded in hardware feedback rather than by one-shot code generation.

\paragraph{Contributions.}
This paper makes the following contributions:
\begin{itemize}

  \item We formulate the problem of \textbf{intent-conditioned CUDA kernel optimization}, mapping either natural-language specifications or existing kernels to performance-optimized CUDA code under correctness and hardware constraints.

  \item We present \textbf{OptiML}, an end-to-end generate-then-optimize framework that unifies a Mixture-of-Thoughts proposal policy generator (\textbf{OptiML-G}) with search-based optimization via Monte Carlo Tree Search (\textbf{OptiML-X}).

  \item We introduce \textbf{OptiML-X}, a CUDA kernel optimizer that explores multi-step program transformations using \textbf{Monte Carlo Tree Search} guided by an \textbf{LLM-as-a-Judge} and profile-derived proxy metrics to target performance bottlenecks.

  \item We provide an \textbf{interpretable evaluation pipeline} based on utilization and work proxies to explain \emph{why} kernels improve, not just \emph{that} they improve, and empirically show that OptiML consistently outperforms strong LLM baselines and optimization-only variants on a diverse suite of CUDA kernels.

\end{itemize}

\section{Related Works}

 Recent work has begun to use LLMs not just to \emph{write} kernels\cite{tehrani2024coderosetta}\cite{pmlr-v162-wen22b}, but to \emph{optimize} them with performance feedback and search. On the evaluation side, \textsc{KernelBench}\cite{ouyang2025kernelbenchllmswriteefficient},\textsc{Kevin}\cite{baronio2025kevinmultiturnrlgenerating} studies whether LLMs can generate efficient PyTorch\cite{paszke2019pytorchimperativestylehighperformance}\emph{GPU kernels} end-to-end, highlighting frequent correctness/compilation failures and a large gap to expert-tuned implementations even when code runs. Complementarily, \textsc{TritonBench}\cite{li2025tritonbenchbenchmarkinglargelanguage} targets the \emph{Triton}\cite{triton} operator domain and provides a controlled benchmark for LLM-generated GPU operators. Moving from benchmarks to optimization systems, several agentic pipelines iteratively propose kernel edits, compile/repair, test, profile, and select improved candidates e.g., \textsc{AutoTriton}\cite{li2025autotritonautomatictritonprogramming} performs automated search/optimization over Triton kernels;  \textsc{Geak}\cite{wang2025geakintroducingtritonkernel} introduces a Triton kernel agent and evaluation benchmarks; and multi-agent frameworks such as \textsc{Astra}\cite{wei2025astramultiagentgpukernel} coordinate diagnosis and rewrite steps for GPU kernel performance.  Other lines of work incorporate hardware-aware signals and structured search in the loop (e.g., \textsc{SwizzlePerf}\cite{tschand2025swizzleperfhardwareawarellmsgpu} for hardware-aware kernel tuning),  and explore CUDA-specific iterative optimization for key kernels like GEMM (e.g., \textsc{CUDA-L2})\cite{su2025cudal2surpassingcublasperformance} or broader evolutionary/iterative improvement settings (e.g., \textsc{EvoEngineer})\cite{guo2025evoengineermasteringautomatedcuda}.  Across these efforts, a common theme is closing the spec$\rightarrow$correct$\rightarrow$fast loop by combining (i) automated compilation repair and test harnesses with (ii) profiling-driven feedback and (iii) explicit search/planning over optimization actions, motivating systems like ours that unify correctness guarding with performance-aware search and LLM judging.

\begin{figure*}[t]
    \centering
    \includegraphics[width=\textwidth]{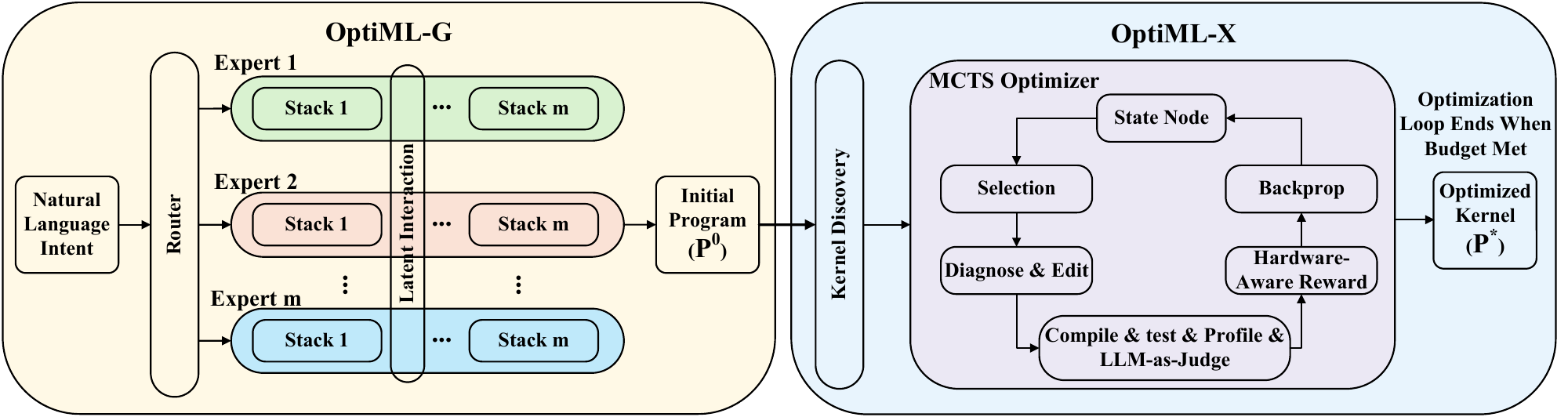}
    \caption{Overview of OptiML. \textbf{OptiML-G} (left) routes inputs through multiple experts with latent-space collaboration to generate initial kernels. \textbf{OptiML-X} (right) optimizes kernels via MCTS with hardware-aware rewards.}
    \label{fig:optiml}
\end{figure*}

\section{\textsc{OptiML}: LLM-Guided CUDA Optimization with Search and Hardware Feedback}
\label{sec:OptiML}
\subsection{OptiML-G}
\label{sec:motgen}

When the input is a natural-language task description rather than existing CUDA code, OptiML first synthesizes an initial kernel via OptiML-G, a code-generation module built on the Mixture of Thoughts (MoT) framework~\cite{feinashley2025mot}. The generated kernel then enters the OptiML-X optimization pipeline (Section~\ref{sec:OptiML-x}).

\subsubsection{Problem Setting}
Given a natural-language specification $\mathcal{D}$ describing the desired kernel functionality, OptiML-G produces an initial CUDA implementation $P_k^{(0)}$ that serves as input to OptiML-X:
\[
\mathcal{D} \;\xrightarrow{\text{OptiML-G}}\; P_k^{(0)} \;\xrightarrow{\text{OptiML-X}}\; P_k^{(*)}.
\]
This end-to-end pipeline enables users to obtain optimized CUDA kernels directly from high-level descriptions without manual implementation.

\subsubsection{Expert Pool and Global Routing}
OptiML-G maintains a pool of $M$ frozen pre-trained code-generation experts $\{\mathcal{E}_0, \dots, \mathcal{E}_{M-1}\}$, each potentially differing in architecture, parameter count, or training corpus. Given input prompt $\mathcal{D}$, a lightweight router $r_\theta$ computes relevance scores over experts:
\[
\mathbf{s} = r_\theta(\mathbf{z}), \quad \mathbf{z} = \textsc{PromptEnc}(\mathcal{D}),
\]
where $\mathbf{z} \in \mathbb{R}^{d}$ is a fixed-dimensional embedding from a frozen encoder. The router selects the top-$K$ experts as the \emph{active set} $\mathcal{I}_{\text{active}} = \textsc{TopK}(\mathbf{s}, K)$, designating the highest-scoring expert as the \emph{primary expert} $m^* = \arg\max_{m \in \mathcal{I}_{\text{active}}} s_m$, which is responsible for final token generation.

\subsubsection{Latent-Space Collaboration via Interaction Layers}
Unlike routing methods that generate from a single expert independently, OptiML-G enables fine-grained collaboration among selected experts in a shared latent space. Each expert $\mathcal{E}_m$ with $L_m$ transformer layers~\cite{vaswani2017attentionneed} is partitioned into $Q$ contiguous stacks:
\[
\mathcal{E}_m = \{S_m^0, S_m^1, \dots, S_m^{Q-1}\}, \quad |S_m^q| = L_m / Q.
\]
At the boundary of each stack $\ell$, the hidden states $\mathbf{h}_m^\ell \in \mathbb{R}^{N \times d_m}$ of active experts are projected into a shared latent dimension $d_s$ via learned projectors:
\[
\mathbf{z}_m^\ell = \textsc{FwdProj}_m^\ell(\mathbf{h}_m^\ell), \quad \mathbf{z}_m^\ell \in \mathbb{R}^{N \times d_s}.
\]
The primary expert $m^*$ then integrates information from all active peers through cross-attention:
\begin{align}
\mathbf{q}^\ell &= \mathbf{z}_{m^*}^\ell, \quad \mathbf{k}^\ell = \mathbf{v}^\ell = \textsc{Concat}\{\mathbf{z}_m^\ell \mid m \in \mathcal{I}_{\text{active}}\}, \\
\mathbf{o}_{m^*}^\ell &= \textsc{CrossAttn}(\mathbf{q}^\ell, \mathbf{k}^\ell, \mathbf{v}^\ell; \mathbf{W}_Q^\ell, \mathbf{W}_K^\ell, \mathbf{W}_V^\ell, \mathbf{W}_O^\ell).
\end{align}
The attended output is projected back and added as a residual update:
\[
\mathbf{h}_{m^*}^\ell \;\leftarrow\; \mathbf{h}_{m^*}^\ell + \textsc{RevProj}_{m^*}^\ell(\mathbf{o}_{m^*}^\ell).
\]
This mechanism allows the primary expert to aggregate complementary ``thoughts'' (hidden representations) from heterogeneous peers at multiple depths, going beyond output-level ensembling while requiring only a single forward pass. Our expert choices are highlighted in Appendix \ref{app:optiml-g}.

\subsubsection{Training Objective}
Only the router $r_\theta$, projectors, and cross-attention weights are trained; all expert backbones remain frozen. The training objective combines language modeling loss with routing regularizers:
\begin{equation}
\mathcal{L}_{\text{total}} = \mathcal{L}_{\text{LM}} + \lambda_{\text{ent}}\mathcal{L}_{\text{ent}} + \lambda_{\text{bal}}\mathcal{L}_{\text{bal}} + \lambda_{\text{con}}\mathcal{L}_{\text{con}},
\label{eq:mot_loss}
\end{equation}
where $\mathcal{L}_{\text{LM}}$ is the autoregressive loss from the primary expert, $\mathcal{L}_{\text{ent}}$ encourages exploration by maximizing router entropy, $\mathcal{L}_{\text{bal}}$ promotes balanced expert utilization, and $\mathcal{L}_{\text{con}}$ enforces routing consistency under stochastic perturbations. Routing gradients are obtained via the Gumbel-Softmax\cite{jang2017categoricalreparameterizationgumbelsoftmax} straight-through estimator. Training in detail is presented in Appendix \ref{app:mot_training}.

\subsubsection{Inference}
At inference time, OptiML-G performs a single forward pass: the router selects active experts, all selected experts process the prompt through their stacks in parallel with interaction layers enabled, and the primary expert emits output tokens while integrating latent information from its peers. This achieves routing-level efficiency with richer representational capacity than single-expert generation.

\subsection{OptiML-X}
\label{sec:OptiML-x}
\subsubsection{Problem Setting}
Given a CUDA source file containing one or more \texttt{\_\_global\_\_} kernels, OptiML-X takes as input \emph{only the CUDA code} (no original prompt or developer annotations) and produces an optimized kernel implementation, along with a structured evaluation report (correctness tests, runtime measurements, and Nsight Compute metrics). Formally, for each discovered kernel $k$ in program $P$, OptiML-X searches over code variants:
\[
P_k^{(0)} \;\rightarrow\; P_k^{(1)} \;\rightarrow\; \cdots \;\rightarrow\; P_k^{(*)},
\]
where $P_k^{(0)}$ is the baseline kernel extracted from the input source, and $P_k^{(*)}$ is the selected optimized kernel variant.

The objective is a constrained optimization:
\begin{align}
\min_{P_k} \quad & T(P_k) \label{eq:objective}\\
\text{s.t.}\quad & \textsc{Correct}(P_k)=1 \label{eq:correctness}\\
& \textsc{Guardrails}(P_k)=1, \label{eq:guardrails}
\end{align}
where $T(P_k)$ is measured runtime (latency), \textsc{Correct} enforces functional correctness, and \textsc{Guardrails} enforces hardware- and compilation-safety constraints derived from profiling signals as shown in \ref{alg:OptiML}.


\subsubsection{Kernel Discovery \& Specifier}
OptiML-X scans the CUDA source and extracts candidate kernels:
\[
\mathcal{K} \;=\; \{k \mid k \text{ is a } \texttt{\_\_global\_\_} \text{ function in the source}\}.
\]
Each kernel $k\in\mathcal{K}$ is optimized independently, enabling multi-kernel compilation units to be processed end-to-end.
Because OptiML-X receives only CUDA code, it must infer the \emph{execution contract} for each kernel: argument kinds (pointer vs.\ scalar), element types, and a plausible workload size (e.g.\ $N$). The Specifier produces a structured specification:
\[
\textsc{Spec}(k) \;=\; \{(a_i, \tau_i, \kappa_i)\}_{i=1}^{m} \cup \textsc{WorkloadParams},
\]
where $a_i$ is an argument, $\tau_i$ is its inferred type, and $\kappa_i\in\{\text{ptr},\text{scalar}\}$ is its kind. This spec is used to synthesize a runner that allocates buffers, initializes inputs, launches $k$, and extracts outputs for correctness checking.

\subsubsection{TestPlan: Multi-Level Correctness}
OptiML-X generates a correctness plan consisting of three tiers:
\begin{itemize}
  \item \textbf{L0 (Smoke):} small sizes and sanity checks to catch crashes, NaNs, and gross shape mismatches.
  \item \textbf{L1 (Randomized):} multiple trials across sizes, seeded randomness, and numerical tolerances.
  \item \textbf{L2 (Metamorphic Relations, optional):} relation tests (e.g.\ scaling invariance, symmetry, permutation equivariance) when the kernel semantics can be inferred confidently.
\end{itemize}
In many kernels, L2 may be empty if no safe metamorphic relation is inferred; OptiML-X still enforces L0/L1 strictly (Eq.~\ref{eq:correctness}).

\subsubsection{Runner Synthesis, Compilation, and Repair}
OptiML-X synthesizes a standalone harness $H_k$ that:
(i) allocates device buffers for pointer arguments,
(ii) initializes inputs (e.g.\ random floats),
(iii)  and launches $k$ under a chosen grid/block configuration,

If compilation fails, OptiML-X enters a bounded repair self refine~\cite{madaan2023selfrefineiterativerefinementselffeedback} loop:
\[
H_k \xrightarrow{\text{nvcc error}} \text{LLM repair} \xrightarrow{} H_k' \xrightarrow{} \cdots
\]
until successful compilation or a maximum repair budget is reached.

\subsubsection{Profiling Signals (Time + NCU)}
For each compiled candidate kernel variant $P_k^{(t)}$, OptiML collects:
\begin{itemize}
  \item \textbf{Runtime:} $T_t$ (mean kernel time / harness time, depending on measurement mode).
  \item \textbf{NCU Metrics:} a vector $\mathbf{m}_t \in \mathbb{R}^d$ extracted from Nsight Compute, including (representative):
  \begin{equation}
  \begin{split}
  \mathbf{m}_t = [\text{SOL}_{sm}, \text{SOL}_{dram}, \text{SOL}_{tex}, \text{dramBytes},\\ \text{l1Sectors}, \text{inst}, \text{warpsActive}, \text{regs}, \dots ].
  \end{split}
  \end{equation}
  
\end{itemize}

\subsubsection{Search-Based Optimization with MCTS}
OptiML-X uses Monte Carlo Tree Search (MCTS) to explore multi-step optimization trajectories (useful when intermediate steps regress performance but enable later improvements). Each node represents a program variant and its measured feedback:
\[
s_t \;=\; (P_k^{(t)},\, T_t,\, \mathbf{m}_t).
\]
Edges correspond to edits proposed by the LLM. MCTS proceeds for a budget $B$ iterations.

\paragraph{Upper Confidence bounds applied to Trees (UCT).}
UCT~\cite{uct} is the selection rule used in Monte Carlo Tree Search (MCTS) to balance
\emph{exploitation} (prefer branches with high observed reward) and
\emph{exploration} (try less-visited branches that may still be promising).
For a parent node $p$ and a child node $c$, UCT selects the child maximizing:
\begin{equation}
\label{eq:uct}
\mathrm{UCT}(c)
= \frac{Q(c)}{N(c)}
+ \alpha \sqrt{\frac{\ln N(p)}{N(c)}} ,
\end{equation}
where $Q(c)$ is the cumulative reward backpropagated to $c$, $N(c)$ is the visit
count of $c$, $N(p)$ is the visit count of $p$, and $\alpha>0$ controls the
exploration strength. The first term $\frac{Q(c)}{N(c)}$ favors children with
high average reward, while the second term encourages exploring children with
high uncertainty (small $N(c)$), ensuring the search does not prematurely
commit to a suboptimal branch.

\paragraph{Diagnosis and Proposal (LLM).}
Given $(P_k^{(t)}, \mathbf{m}_t)$, the LLM proposes a ranked list of bottleneck hypotheses (e.g.\ compute-bound, cache-bound, occupancy-limited), then emits a concrete patch/edit addressing the selected hypothesis (e.g.\ tiling, vectorization, unrolling, launch changes).

\paragraph{Execution.}
Each proposal is materialized, compiled, and evaluated:
\[
P_k^{(t)} \xrightarrow{\text{patch}} P_k^{(t+1)} \xrightarrow{\text{compile}} \xrightarrow{\text{tests}} \xrightarrow{\text{profile}} (T_{t+1}, \mathbf{m}_{t+1}).
\]
Candidates failing compilation or L0/L1 tests receive strong negative reward and are pruned.

\subsubsection{Reward Model: Time + Proxy Metrics + Guardrails + LLM Judge}
OptiML-X uses a composite reward that blends \emph{measured runtime}, \emph{hardware proxy signals}, \emph{guardrails}, and an \emph{LLM judge}. Let baseline be $(T_0, \mathbf{m}_0)$ and candidate be $(T, \mathbf{m})$.

\paragraph{Time Reward.}
We define a bounded time-improvement term:
\begin{equation}
r_{\text{time}} \;=\; \tanh\!\left(\alpha_t \cdot \frac{T_0 - T}{\max(T_0,\epsilon)}\right),
\label{eq:time_reward}
\end{equation}
where $\alpha_t$ controls sensitivity.

\paragraph{Proxy Reward from NCU Metrics.}
Let $\Delta^{+}\SOL_m = \SOL_m-\SOL_{m,0}$ denote an increase in utilization (good),
and $\Delta^{-}X_x = X_{x,0}-X_x$ denote a reduction in work metrics (good).
We define the shaping reward:
\begin{equation}\scriptsize
\begin{split} 
\label{eq:proxy_reward}
r_{\text{proxy}}=
\sum_{m\in\{sm,dram,tex\}} w_m\,\clip\!\Big(\tfrac{\Delta^{+}\SOL_m}{100}\Big)\\
+\sum_{x\in\{\texttt{l1\_sectors},\texttt{dram\_bytes},\texttt{inst\_executed}\}}
w_x\,\clip\!\Big(\tfrac{\Delta^{-}X_x}{X_{x,0}+\epsilon}\Big),
\end{split}
\end{equation}
where $\clip(z)=\max(-1,\min(1,z))$ and $\epsilon$ prevents division by zero.
This proxy is a shaping signal (not a substitute for time) that favors edits which
increase utilization and/or reduce memory traffic and instruction footprint.
\paragraph{Bottleneck-aware reward.}
We use bottleneck-aware proxy weighting: the reward reweights utilization/traffic terms depending on whether the kernel is inferred to be compute- or memory-bound from SOL gaps (details in Appendix~\ref{app:bottleneck_weights}).
\paragraph{ Guardrails (hard penalties).}
We enforce penalties for pathological candidates:
\begin{equation}
\begin{split}  
p_{\text{guard}} \;=\; 
\lambda_{\text{spill}}\cdot \mathbb{I}[\text{lmemBytes} > 0]
\;+\;
\lambda_{\text{regs}}\cdot \mathbb{I}[\text{regs} > \rho_{\max}]\\
\;+\;
\lambda_{\text{work}}\cdot \mathbb{I}[\text{dramBytes} \uparrow\!\!\uparrow \;\vee\; \text{inst} \uparrow\!\!\uparrow],
\label{eq:guardrails_penalty}
\end{split}
\end{equation}
where $\uparrow\!\!\uparrow$ denotes exceeding a configured inflation threshold relative to baseline (e.g.\ $>1.5\times$).

\paragraph{LLM-as-a-Judge.}
The judge receives baseline vs.\ candidate summaries, including $(T_0,\mathbf{m}_0)$ and $(T,\mathbf{m})$ and the edit description, and outputs $r_{\text{llm}} \in [-1,1]$ and $v \in \{\text{KEEP},\text{DISCARD}\}$, where $r_{\text{llm}}$ evaluates whether the edit plausibly addresses the diagnosed bottleneck and whether the measured signals corroborate that claim.

\paragraph{Final Composite Reward.}
The total reward is:
\begin{equation}
R \;=\; \beta_{\text{time}}\, r_{\text{time}} \;+\; \beta_{\text{proxy}}\, r_{\text{proxy}} \;+\; \beta_{\text{llm}}\, r_{\text{llm}} \;-\; p_{\text{guard}},
\label{eq:total_reward}
\end{equation}
with $\beta$ weights chosen to keep $r_{\text{time}}$ dominant while allowing proxy shaping and judge consistency. If L0/L1 fails, OptiML-X assigns $R=-\infty$ (or a large negative constant) and rejects the candidate.

\subsection{Backpropagation and Final Selection}
After evaluating a candidate with reward $R$, MCTS backpropagates the reward along the selected path, updating visit counts and cumulative rewards used by UCT (Eq.~\ref{eq:uct}). OptiML then selects the final output as \textbf{Best KEEP chain:} among candidates labeled \texttt{KEEP}, choose the one with highest $R$. and \textbf{Fallback:} if no \texttt{KEEP} exists, return the baseline. When keep-chaining is enabled, the final code is the result of applying the sequence of \texttt{KEEP} edits along the chosen path (not merely the best single edit in isolation).


\section{Evaluation Harness}
\label{sec:eval_harness}

\textbf{Hardware \& measurement.}
All experiments run on a NVIDIA A100 80\,GB GPU. Kernels are compiled with \texttt{nvcc} (CUDA~12.6) and timed with CUDA events (2 warm-up + 10 timed iterations). Nsight Compute is run in separate profiling passes to collect SOL utilization metrics and instruction/traffic counters under a fixed metric set. \\
\textbf{OptiML-X} uses \textbf{GPT-5 mini}\cite{singh2025openaigpt5card} internally.

\textbf{Optimization pipeline and budgets.}
For each kernel, LLM baselines are first compilation-gated via an automated self-repair loop (up to $B_{\text{repair}}{=}3$ attempts); candidates still failing are marked \texttt{CF}. OptiML-X performs MCTS over semantics-preserving code edits (tiling, shared-memory staging, vectorized/coalesced memory ops, unrolling, warp-level primitives) with  maximum depth $d_{\max}{=}6$, and UCT exploration $c_{\text{uct}}{=}1.4$. An LLM-as-a-Judge provides an edit prior and legality/bottleneck critiques using only code and previously observed profiling counters.\\


\section{Results}

\begin{table*}[!ht]
\centering
\resizebox{\textwidth}{!}{%
\begin{tabular}{|c|c|c|ccc|ccc|}
\hline
\multicolumn{3}{|c|}{} & \multicolumn{3}{c|}{\textbf{Hardware Utilization SOL $\uparrow$}} & \multicolumn{3}{c|}{\textbf{Work Done $\downarrow$}} \\ \hline
\textbf{Task} & \textbf{Model} & \textbf{Time (speedup)} &
\textbf{Compute sm} & \textbf{Memory dram} & \textbf{Texture tex} &
\textbf{dram\_bytes} & \textbf{l1\_sectors} & \textbf{inst\_executed} \\ \hline

\multirow{9}{*}{Matrix Multiplication}
& GPT 5.1
& 7.20
& 71.26 & 8.42 & 91.36
& \best{63} & 408 & 419 \\
& Qwen2.5-Coder
& 7.65 (0.94$\times$)
& 69.40 & 8.10 & 90.98
& \second{66} & 421 & 432 \\
& HPC-Coder-V2
& 7.88 (0.91$\times$)
& 68.72 & 8.02 & 90.71
& \third{67} & 427 & 438 \\
& StarCoder2
& 8.23 (0.87$\times$)
& 67.95 & 7.92 & 90.41
& 68 & 435 & 441 \\
& GPT 5.1 + OptiML-X
& \third{4.62} (1.56$\times$)
& \third{78.41} & \third{10.62} & \third{92.07}
& 79 & \third{292} & \third{392} \\
& Qwen2.5-Coder + OptiML-X
& 4.98 (1.45$\times$)
& 76.58 & 10.14 & 91.22
& 74 & 318 & 400 \\
& HPC-Coder-V2 + OptiML-X
& \second{4.52} (1.59$\times$)
& \second{79.16} & \second{10.86} & \second{92.21}
& 73 & \second{286} & \second{386} \\
& StarCoder2 + OptiML-X
& 4.74 (1.52$\times$)
& 77.42 & 10.33 & 91.63
& 77 & 301 & 395 \\
& OptiML
& \best{4.40} (1.64$\times$)
& \best{81.17} & \best{11.03} & \best{92.61}
& 70 & \best{282} & \best{381} \\ \hline

\multirow{9}{*}{Max Pooling 3D}
& GPT 5.1 & CF & -- & -- & -- & -- & -- & -- \\
& Qwen2.5-Coder & CF & -- & -- & -- & -- & -- & -- \\
& HPC-Coder-V2 & CF & -- & -- & -- & -- & -- & -- \\
& StarCoder2 & CF & -- & -- & -- & -- & -- & -- \\
& GPT 5.1 + OptiML-X
& 6.09
& \third{28.14} & \second{31.47} & \third{36.72}
& \third{524} & \third{1944} & \third{824} \\
& Qwen2.5-Coder + OptiML-X
& 6.38 (0.95$\times$)
& 26.88 & \best{32.12} & 35.31
& 548 & 2032 & 843 \\
& HPC-Coder-V2 + OptiML-X
& \second{5.88} (1.04$\times$)
& \second{29.02} & 30.41 & \second{37.11}
& \second{508} & \second{1891} & \second{809} \\
& StarCoder2 + OptiML-X
& \third{6.02} (1.01$\times$)
& 27.61 & \third{31.18} & 36.02
& 532 & 1978 & 831 \\
& OptiML
& \best{5.62} (1.08$\times$)
& \best{30.71} & 29.18 & \best{37.55}
& \best{494} & \best{1863} & \best{801} \\ \hline

\multirow{9}{*}{Multi-Head Self-Attention}
& GPT 5.1 & CF & -- & -- & -- & -- & -- & -- \\
& Qwen2.5-Coder & CF & -- & -- & -- & -- & -- & -- \\
& HPC-Coder-V2 & CF & -- & -- & -- & -- & -- & -- \\
& StarCoder2 & CF & -- & -- & -- & -- & -- & -- \\
& GPT 5.1 + OptiML-X
& \third{7.90}
& \third{48.62} & \third{26.41} & \third{60.32}
& \third{1117} & \third{4213} & \third{2094} \\
& Qwen2.5-Coder + OptiML-X
& 9.10 (0.87$\times$)
& 46.37 & \best{27.91} & 58.24
& 1183 & 4368 & 2172 \\
& HPC-Coder-V2 + OptiML-X
& \second{7.55} (1.05$\times$)
& \second{49.31} & 25.72 & \second{60.88}
& \second{1086} & \second{4124} & \second{2048} \\
& StarCoder2 + OptiML-X
& 8.22 (0.96$\times$)
& 47.55 & \second{26.98} & 59.46
& 1144 & 4286 & 2126 \\
& OptiML
& \best{6.85} (1.15$\times$)
& \best{50.28} & 24.88 & \best{61.79}
& \best{1064} & \best{3961} & \best{2012} \\ \hline

\multirow{9}{*}{ReLU Activation Function}
& GPT 5.1
& 7.50
& 12.11 & 41.37 & 16.44
& 263 & 883 & 358 \\
& Qwen2.5-Coder
& 7.80 (0.96$\times$)
& 11.74 & \third{42.12} & 16.06
& 274 & 918 & 372 \\
& HPC-Coder-V2
& 8.02 (0.93$\times$)
& 11.48 & \second{42.43} & 15.89
& 281 & 939 & 381 \\
& StarCoder2
& 8.31 (0.90$\times$)
& 11.21 & \best{42.77} & 15.73
& 289 & 962 & 392 \\
& GPT 5.1 + OptiML-X
& \third{4.87} (1.54$\times$)
& \third{15.64} & 34.28 & \third{18.71}
& \third{212} & \third{717} & \third{332} \\
& Qwen2.5-Coder + OptiML-X
& 5.24 (1.43$\times$)
& 14.92 & 35.14 & 18.21
& 220 & 751 & 342 \\
& HPC-Coder-V2 + OptiML-X
& \second{4.68} (1.60$\times$)
& \second{15.98} & 33.61 & \second{18.92}
& \second{206} & \second{701} & \second{327} \\
& StarCoder2 + OptiML-X
& 4.99 (1.50$\times$)
& 15.12 & 34.72 & 18.44
& 214 & 726 & 335 \\
& OptiML
& \best{4.49} (1.67$\times$)
& \best{16.72} & 32.03 & \best{19.37}
& \best{201} & \best{689} & \best{322} \\ \hline

\multirow{9}{*}{Categorical Cross-Entropy Loss}
& GPT 5.1 & CF & -- & -- & -- & -- & -- & -- \\
& Qwen2.5-Coder & CF & -- & -- & -- & -- & -- & -- \\
& HPC-Coder-V2 & CF & -- & -- & -- & -- & -- & -- \\
& StarCoder2 & CF & -- & -- & -- & -- & -- & -- \\
& GPT 5.1 + OptiML-X
& \third{6.90}
& \third{36.22} & \third{27.41} & \third{30.18}
& \third{944} & \third{2412} & \third{1713} \\
& Qwen2.5-Coder + OptiML-X
& 7.35 (0.94$\times$)
& 34.66 & \best{28.62} & 29.04
& 979 & 2514 & 1809 \\
& HPC-Coder-V2 + OptiML-X
& \second{6.62} (1.04$\times$)
& \second{36.88} & \second{26.44} & \second{30.74}
& \second{917} & \second{2346} & \second{1681} \\
& StarCoder2 + OptiML-X
& 7.08 (0.97$\times$)
& 35.14 & 27.86 & 29.62
& 956 & 2451 & 1732 \\
& \textsc{OptiML}
& \best{6.40} (1.08$\times$)
& \best{37.94} & 25.31 & \best{31.46}
& \best{900} & \best{2272} & \best{1657} \\ \hline

\end{tabular}%
}
\caption{\textbf{End-to-end latency and Nsight Compute characterization} on five representative CUDA-LLM kernels from the task suite~\cite{chen2025cudallmllmswriteefficient} (A100 80GB). We report runtime (in seconds) (lower is better) together with utilization proxies (SM/DRAM/TEX SOL; higher is better) and work proxies (DRAM bytes, L1 sectors, and executed instructions; lower is better). For each task, we compare LLM-only baselines, their OptiML-X post-optimization variants, and \textbf{OptiML} (G+X). Many LLM-only baselines fail to compile (CF). \best{Best}, \second{second}, \third{third} indicate per-column ranks within each task block.}
\label{tab:sub_tasks_results}
\end{table*}

\subsection{Bottlenecks and how OptiML-X (MCTS + LLM-as-a-Judge) alleviates them.}
Beyond raw speed, the Nsight metrics in Table~\ref{tab:sub_tasks_results} expose the dominant bottleneck class for each kernel and clarify \emph{why} OptiML-X improves performance.
\subsubsection{Matrix Multiplication}
For \textit{Matrix Multiplication}, the baseline already exhibits high \texttt{tex\_SOL} ($\sim$91--92) and moderate \texttt{sm\_SOL} ($\sim$68--71), indicating a compute-/throughput-limited kernel where the primary headroom comes from improving instruction efficiency and memory locality rather than reducing total bytes. OptiML-X increases \texttt{sm\_SOL}/\texttt{dram\_SOL}/\texttt{tex\_SOL} (e.g., GPT~5.1: 71.26/8.42/91.36 $\rightarrow$ OptiML: 81.17/11.03/92.61) while simultaneously reducing \texttt{l1\_sectors} and \texttt{inst\_executed} (408$\rightarrow$282 and 419$\rightarrow$381), consistent with better tiling, coalescing, and fewer redundant address/instruction sequences (e.g., unrolling choices, register reuse, and shared-memory staging). Our judge-guided MCTS favors rewrites that increase utilization while reducing instruction and L1 transaction counts, selecting transformations that approach peak throughput without sacrificing correctness.

\subsubsection{Max Pooling 3D}
For \textit{Max Pooling 3D}, the comparatively low \texttt{sm\_SOL} (26.9--30.7) coupled with substantially higher \texttt{dram\_SOL} (29--32) and large \texttt{l1\_sectors} (1863--2032) points to a memory-traffic bottleneck driven by poor reuse and scattered access patterns typical of 3D windowing. OptiML-X reduces work metrics (\texttt{dram\_bytes}: 524$\rightarrow$494, \texttt{l1\_sectors}: 1944$\rightarrow$1863, \texttt{inst\_executed}: 824$\rightarrow$801) while increasing \texttt{sm\_SOL} and \texttt{tex\_SOL} (28.14/36.72 $\rightarrow$ 30.71/37.55), suggesting improved data reuse and fewer redundant loads (e.g., staging tiles in shared memory, fusing boundary checks, and reducing repeated global reads within the pooling window). Here, the judge-guided MCTS is crucial: among many plausible rewrites, it prioritizes those that measurably lower L1 sector traffic (a proxy for cache-line transactions) while preserving or increasing SM throughput.

\subsubsection{Multi-Head Self-Attention}
For \textit{Multi-Head Self-Attention}, the high instruction count and large L1 traffic (\texttt{inst\_executed} $\approx$2012--2172; \texttt{l1\_sectors} $\approx$3961--4368) alongside mid-to-high utilization (\texttt{sm\_SOL} $\approx$46--50, \texttt{tex\_SOL} $\approx$58--62) indicate a mixed bottleneck: the kernel is both instruction-heavy (softmax/normalization, address arithmetic) and memory-intensive (Q/K/V loads, intermediate reductions). OptiML-X reduces \texttt{dram\_bytes}/\texttt{l1\_sectors}/\texttt{inst\_executed} while increasing \texttt{sm\_SOL} and \texttt{tex\_SOL} (e.g., GPT~5.1+OptiML-X: 1117/4213/2094 $\rightarrow$ OptiML: 1064/3961/2012 and 48.62/60.32 $\rightarrow$ 50.28/61.79), consistent with more effective fusion and better loop structure (e.g., fewer separate passes over the same data, improved vectorized loads, and reduced control divergence). Importantly, the OptiML-X runtimes vary across generators (7.55--9.10\,s), illustrating why MCTS matters: the judge provides a dense signal over intermediate candidates, enabling the search to escape local optima (e.g., ``faster'' rewrites that accidentally increase instruction count or L1 transactions) and converge toward globally better utilization/work trade-offs.

\subsubsection{ReLU Activation Function}
For \textit{ReLU}, the baseline behavior is dominated by overhead rather than heavy computation: \texttt{sm\_SOL} and \texttt{tex\_SOL} are comparatively low (11--12 and 15--16 for LLM-only baselines), while the work counters remain nontrivial for such a simple elementwise operation, indicating inefficiencies such as uncoalesced accesses, suboptimal vectorization, and extra instructions for bounds/strides. OptiML-X improves utilization and reduces work (e.g., GPT~5.1+OptiML-X: \texttt{l1\_sectors}/\texttt{inst\_executed}=717/332 $\rightarrow$ OptiML: 689/322 with higher \texttt{sm\_SOL}/\texttt{tex\_SOL}=16.72/19.37), aligning with transformations like vectorized loads/stores (e.g., \texttt{float4}), reduced index arithmetic, and simplified control flow. The judge-guided MCTS is effective here because small code changes yield measurable differences in instruction count and cache behavior; the search selects candidates that reduce overhead while preserving correctness.

\subsubsection{Categorical Cross-Entropy Loss}
Finally, for \textit{Categorical Cross-Entropy}, the dominant bottleneck is the combination of reduction-like control flow and memory traffic: mid-range \texttt{sm\_SOL} (34--38) with sizeable work counters (\texttt{l1\_sectors} 2272--2514; \texttt{inst\_executed} 1657--1809) suggests that performance is constrained by reduction structure (e.g., partial sums, log/exp operations, and synchronization patterns) as much as by raw memory. OptiML-X reduces \texttt{dram\_bytes}/\texttt{l1\_sectors}/\texttt{inst\_executed} (e.g., 944/2412/1713 $\rightarrow$ 900/2272/1657) while increasing \texttt{sm\_SOL} and \texttt{tex\_SOL} (36.22/30.18 $\rightarrow$ 37.94/31.46), consistent with improved reduction strategy (e.g., warp-level primitives, fewer synchronizations, better batching of arithmetic). Since all LLM-only baselines fail to compile (CF), this kernel also highlights the second role of our framework: the judge penalizes candidates that violate compilation constraints early in the search, increasing the probability of reaching a valid, high-performance implementation.

Overall, the per-kernel analysis shows that OptiML-X does not apply a single heuristic; instead, MCTS steers exploration using the LLM-as-a-Judge to optimize for the \emph{observed} limiting factor: decreasing L1 transactions and redundant loads for memory-bound kernels (pooling), reducing instruction and control overhead for elementwise kernels (ReLU), and jointly improving fusion/tiling to cut both instruction count and cache traffic in mixed kernels (attention, cross-entropy), which is reflected quantitatively by higher achieved utilization and lower work counters alongside consistent end-to-end speedups.

\begin{table}[!ht]
\centering
\caption{Performance comparison on ParEval benchmark (\%).}
\label{tab:pareval}
\resizebox{\columnwidth}{!}{%
\begin{tabular}{lccc}
\toprule
\textbf{Model} & \textbf{Pass@1} & \textbf{Pass@5} & \textbf{Pass@10} \\
\midrule
Qwen2.5-Coder-7B & 2.33 & 3.33 & 3.33 \\
StarCoder2-7B & 2.67 & 6.84 & 8.79 \\
HPC-Coder-V2-6.7B & 4.74 & 10.23 & 14.21 \\
GPT-5.1 & 10.46 & 24.71 & 32.26 \\
\midrule
OptiML-G (Ours) & \textbf{12.17} & \textbf{27.08} & \textbf{33.33} \\
\bottomrule
\end{tabular}%
}
\end{table}

\subsection{Ablation Study.}
\subsubsection{Code Generation Quality}
To evaluate the effectiveness of OptiML-G as a standalone code generation module, we assess its performance on the ParEval\cite{nichols2024large} benchmark, which measures the ability to generate correct parallel CUDA code from natural-language specifications. Table~\ref{tab:pareval} reports Pass@$k$ metrics, where Pass@$k$ denotes the probability that at least one of $k$ generated samples passes all correctness tests. OptiML-G achieves 12.17\% Pass@1, 27.08\% Pass@5, and 33.33\% Pass@10, consistently outperforming all baseline models across all metrics. Notably, OptiML-G improves Pass@1 by 1.71 percentage points over GPT-5.1 (10.46\%), demonstrating that the Mixture-of-Thoughts (MoT) framework enables more reliable first-attempt code generation. The gains are more pronounced at higher $k$ values, with OptiML-G achieving a 2.37-point improvement over GPT-5.1 at Pass@5 and a 1.07-point improvement at Pass@10. Compared to specialized code models such as HPC-Coder-V2 (4.74\% Pass@1) and StarCoder2 (2.67\% Pass@1), OptiML-G exhibits substantially stronger performance, indicating that latent-space collaboration among heterogeneous experts produces structurally superior kernels. 
\subsubsection{Code Optimization Emphasis}
\label{sec:ablation_pass_speedup}
\begin{table}[!ht]
\centering
\caption{Pass@\{1,5,10\} and Speedup@\{1,5,10\} on the 5-task subset under best-of-$k$ selection. Speedup is geometric mean over solved tasks with per-task baseline: GPT~5.1 when available, else GPT~5.1+OptiML-X when GPT~5.1 is CF.}
\label{tab:pass_speedup_atk_subset}
\resizebox{\columnwidth}{!}{%
\begin{tabular}{lccc|ccc}
\toprule
\textbf{Model} &
\textbf{Pass@1} & \textbf{Pass@5} & \textbf{Pass@10} &
\textbf{Speedup@1} & \textbf{Speedup@5} & \textbf{Speedup@10} \\
\midrule
Qwen2.5-Coder            & 20.0 & 40.0 & 40.0 & 0.89 & 0.93 & 0.95 \\
StarCoder2               & 20.0 & 40.0 & 40.0 & 0.82 & 0.86 & 0.89 \\
HPC-Coder-V2             & 20.0 & 40.0 & 40.0 & 0.86 & 0.90 & 0.92 \\
GPT 5.1                  & 40.0 & 40.0 & 40.0 & 1.00 & 1.00 & 1.00 \\
\midrule
GPT 5.1 + OptiML-X        & 80.0 & 100.0 & 100.0 & 1.10 & 1.16 & 1.19 \\
Qwen2.5-Coder + OptiML-X  & 70.0 & 100.0 & 100.0 & 1.03 & 1.08 & 1.10 \\
HPC-Coder-V2 + OptiML-X   & 85.0 & 100.0 & 100.0 & 1.14 & 1.21 & 1.24 \\
StarCoder2 + OptiML-X     & 75.0 & 100.0 & 100.0 & 1.06 & 1.12 & 1.17 \\
\midrule
OptiML (Gen + X)          & \textbf{85.0} & \textbf{100.0} & \textbf{100.0} &
\textbf{1.18} & \textbf{1.26} & \textbf{1.30} \\
\bottomrule
\end{tabular}%
}
\end{table}
Table~\ref{tab:pass_speedup_atk_subset} separates the effects of sampling, post-hoc optimization, and the full OptiML pipeline under best-of-$k$. LLM-only baselines have low success (Pass@1 = 20\%--40\%) and saturate at Pass@10 = 40\%, with Speedup@10 $\le 1.00$, indicating that additional samples rarely fix compilation/performance issues. Adding OptiML-X sharply improves reliability (Pass@1 = 70\%--85\%, Pass@5 = 100\%) and consistently increases speedup (e.g., HPC-Coder-V2+OptiML-X: 1.14/1.21/1.24 at $k=\{1,5,10\}$). The full OptiML (Gen+X) matches the best Pass@1 (85\%) while achieving the highest speedups at all $k$ (1.18/1.26/1.30), showing that OptiML-Gen provides stronger starting kernels that OptiML-X can optimize more effectively.\\
Additional OptiML-X details (prompt/tool schemas, judge + composite reward/guardrails, and the MCTS-budget ablations) are provided in Appendix~B--C and Appendix~E, with complete per-kernel results deferred to Appendix~D.
\section{Conclusion}
We presented \textbf{OptiML}, an end-to-end framework for \emph{LLM-driven CUDA kernel development} that combines \textbf{OptiML-G} (mixture-of-thoughts code generation to improve compilability and structural quality) with \textbf{OptiML-X} (an MCTS-based test-time optimizer guided by profiling signals and an LLM-as-a-Judge). Across a diverse CUDA kernel suite on A100, OptiML consistently improves end-to-end runtime over strong LLM-only baselines and over post-hoc optimization alone. Importantly, the gains are not merely from ``trying more variants'': OptiML-X uses profile-derived utilization and work proxies to target bottlenecks (e.g., memory traffic and instruction footprint) while avoiding edits that break correctness or trade one limit for another, yielding a more sample-efficient and interpretable search process.

\section{Acknowledgments}
This research was supported by the National Science Foundation under grant number 2211982. We would also like to thank Cisco AI Research for their generous support.

\bibliography{example_paper}
\bibliographystyle{icml2026}

\newpage
\appendix
\onecolumn
\section{OptiML-X Algorithm}
\begin{algorithm}[!ht]
\caption{OptiML-X per-kernel optimization}
\label{alg:OptiML}
\begin{algorithmic}[1]\scriptsize
\REQUIRE CUDA source $P$, kernel name $k$, search budget $B$, GPU $sm\_XX$
\STATE $\textsc{Spec}\gets \textsc{Specifier}_{\text{LLM}}(P,k)$
\STATE $\textsc{Plan}\gets \textsc{TestPlan}_{\text{LLM}}(P,k,\textsc{Spec})$
\STATE $H \gets \textsc{SynthesizeRunner}(P,k,\textsc{Spec})$
\STATE $(ok,H)\gets \textsc{CompileWithRepair}(H)$
\IF{$\neg ok$} \STATE \texttt{$compile\_failed$} \ENDIF
\STATE $\textsc{Assert}(\textsc{RunTests}(H,\textsc{Plan}) \text{ passes L0/L1})$
\STATE $(T_0,\mathbf{m}_0)\gets \textsc{Profile}(H)$
\STATE Initialize MCTS tree root $s_0=(P,T_0,\mathbf{m}_0)$
\FOR{$i=1$ to $B$}
  \STATE Select leaf $s$ by UCT
  \STATE $\mathcal{H}\gets \textsc{Diagnose}_{\text{LLM}}(s.P, s.\mathbf{m})$
  \STATE $h \gets \textsc{PickHypothesis}(\mathcal{H})$
  \STATE $\textsc{patch}\gets \textsc{Propose}_{\text{LLM}}(s.P,h,s.\mathbf{m})$
  \STATE $P'\gets \textsc{ApplyPatch}(s.P,\textsc{patch})$
  \STATE $H'\gets \textsc{SynthesizeRunner}(P',k,\textsc{Spec})$
  \IF{$\neg \textsc{Compile}(H')$ or $\neg \textsc{PassL0L1}(H',\textsc{Plan})$}
    \STATE Backpropagate large negative reward; \textbf{continue}
  \ENDIF
  \STATE $(T,\mathbf{m})\gets \textsc{Profile}(H')$
  \STATE $(r_{\text{llm}}, v)\gets \textsc{Judge}_{\text{LLM}}((T_0,\mathbf{m}_0),(T,\mathbf{m}),\textsc{patch})$
  \STATE $R \gets \textsc{CompositeReward}(T_0,\mathbf{m}_0,T,\mathbf{m},r_{\text{llm}})$
  \STATE Add child node and backpropagate $R$
\ENDFOR
\STATE best candidate among \texttt{KEEP} nodes (else baseline), plus reports
\end{algorithmic}
\end{algorithm}

\section{Training OptiML-G (Mixture-of-Thoughts Router)}
\label{app:mot_training}

\paragraph{Overview.}
OptiML-Gen is implemented as a \emph{mixture-of-thoughts} (MoT) router over a fixed set of CUDA-capable code LLM experts. During training we \emph{freeze all expert weights} and learn only lightweight \textbf{routing} and \textbf{interaction} modules that (i) select which experts to query per input and (ii) fuse their intermediate representations into a single next-token distribution. This keeps training stable and inexpensive while preserving each expert's specialization.

\paragraph{Experts and freezing.}
We use three decoder-only experts: \texttt{Qwen2.5-Coder-7B}, \texttt{HPC-Coder-V2-6.7B}, and \texttt{StarCoder2-7B}. All expert parameters are set \texttt{requires\_grad=False}; only the MoT router + interaction layers are optimized. Experts are loaded in FP16 by default (quantization flags exist but are disabled in our setup), and we use a shared tokenizer/padding strategy (pad token set to EOS when missing).

\paragraph{Training data and formatting.}
Training uses instruction-style supervision with 6033 paired \texttt{cpp\_code}, \texttt{cuda\_code}, and a natural-language \texttt{description}. Each sample is rendered into a single causal-LM sequence using a fixed template:
\emph{Task description} $\rightarrow$ \emph{C++ input} $\rightarrow$ \emph{CUDA output}.
We train with standard next-token prediction on the full sequence (teacher forcing). Sequences are truncated/padded to a maximum length (we use 512 in our runs).

\paragraph{MoT architecture and routing.}
We stack $S$ MoT blocks (default $S{=}4$). In each block, the router produces expert scores and selects \texttt{top-k} experts per token (typically $k{=}2$) using a stochastic Gumbel-Softmax gate during training (temperature $\tau$). Selected expert hidden states are projected into a shared space of dimension $d$ (e.g., $d{=}768$) and fused using a multi-head attention \emph{interaction module} (e.g., 8 heads). The fused representation is passed forward to the next MoT block and ultimately to the LM head for logits.

\paragraph{Objective and regularization.}
The primary objective is standard cross-entropy language modeling loss. We add two auxiliary terms that improve routing behavior:
(i) an \textbf{expert load-balancing} loss to discourage collapse onto a single expert, and
(ii) a \textbf{consistency} loss that encourages routing stability by comparing predictions across two stochastic forward passes (different Gumbel noise) on the same batch.
Concretely, we minimize:
\[
\mathcal{L} \;=\; \mathcal{L}_{\text{LM}} \;+\; \lambda_{\text{cons}}\mathcal{L}_{\text{cons}} \;+\; \lambda_{\text{bal}}\mathcal{L}_{\text{bal}},
\]
with $\lambda_{\text{cons}}{=}0.05$ in our runs; load-balancing/entropy regularization are applied with small fixed weights in the implementation to keep routing diverse without dominating $\mathcal{L}_{\text{LM}}$.

\paragraph{Optimization details.}
We train with AdamW on the router/interaction parameters only (expert params excluded), cosine warm restarts schedule, FP16 enabled, and gradient accumulation to reach an effective batch size $>1$ despite long sequences. A representative configuration (matching our training script/README) is:
\begin{itemize}\itemsep2pt
\item epochs: 3,\;\; lr: $5\times10^{-6}$,\;\; warmup steps: 200,\;\; max length: 512
\item per-device batch size: 1,\;\; grad accumulation: 8,\;\; grad clip: 0.3
\item MoT: num\_stacks $=4$, top\_k $=2$, shared\_dim $=768$, router hidden $=512$, heads $=8$
\end{itemize}
We checkpoint the best model on validation loss and use it for OptiML-G inference (best-of-$k$ generation over multiple routed samples) before handing candidates to OptiML-X.

\section{Prompts}
Table~\ref{tab:cudallm_task_prompts} provides a descriptive prompt which is used to generate code for each task.

\begin{table*}[!ht]
\centering
\caption{\textbf{Benchmark task prompts used for initial prompt construction (Appendix C / Table 4).}
Each row provides the high-level natural-language task specification injected into the initial prompt template.}
\label{tab:cudallm_task_prompts}
\scriptsize
\renewcommand{\arraystretch}{1.15}
\setlength{\tabcolsep}{6pt}
\begin{tabular}{c l p{0.70\textwidth}}
\toprule
\textbf{Task ID} & \textbf{Task} & \textbf{Prompt} \\
\midrule

1 & Sigmoid &
Implement a CUDA program for sigmoid activation function: $\mathrm{sigmoid}(x)=\frac{1}{1+\exp(-x)}$.
Input shape: (batch size, dim); Output: same shape as input. \\
\midrule
2 & Matrix Multiplication &
Write a program that multiplies two matrices of 32-bit floating point numbers on a GPU.
Given matrix $A$ of dimensions $M \times K$ and matrix $B$ of dimensions $K \times N$,
compute the product matrix $C = A \times B$, which will have dimensions $M \times N$. \\
\midrule
3 & Max Pooling 3D &
Implement a CUDA program for 3D max pooling function that selects the maximum value within a defined region
(a window) of a feature map. Input shape: (batch size, channels, dim1, dim2, dim3);
Output: with 3D max pooling applied. \\
\midrule
4 & LayerNorm &
Implement a GPU program that performs Layer Normalization (LayerNorm) operation, which normalizes across the
features for each individual data sample in a layer. Input of shape (batch size, features, dim1, dim2);
Output with Layer Normalization applied, same shape as input. \\
\midrule
5 & 2D Convolution &
Write a program that performs a 2D convolution operation on the GPU. Given an input matrix and a kernel (filter),
compute the convolved output. The convolution should be performed with a ``valid'' boundary condition, meaning
the kernel is only applied where it fully overlaps with the input.
The input consists of:
(1) input: A 2D matrix of 32-bit floating-point numbers, represented as a 1D array in row-major order.
(2) kernel: A 2D kernel (filter) of 32-bit floating-point numbers, also represented as a 1D array in row-major order.
The output should be written to the output matrix (also a 1D array in row-major order).
The output matrix will have dimensions:
output rows = input rows - kernel rows + 1,
output cols = input cols - kernel cols + 1.
The convolution operation is defined as:
output[i][j] = $\sum_{m=0}^{\mathrm{kernel\ rows}-1}\sum_{n=0}^{\mathrm{kernel\ cols}-1}$ input[i+m][j+n] * kernel[m][n]. \\
\midrule
6 & Multi-Head Self-Attention &
Implement a CUDA program for multi-head self-attention. Given three input matrices $Q$ (queries), $K$ (keys),
and $V$ (values) of size $N \times d_{\text{model}}$, compute:
$\mathrm{MultiHead}(Q,K,V)=\mathrm{Concat}(head_1,\ldots,head_h)$,
where each head computes:
$head_i=\mathrm{softmax}\!\left(\frac{Q_i K_i^{T}}{\sqrt{d_k}}\right)V_i$
with $d_k=d_{\text{model}}/h$ and $Q_i,K_i,V_i$ being the $i$-th head’s partition of the input matrices. \\
\midrule
7 & Mean Square Error &
Implement a CUDA program to calculate the Mean Squared Error (MSE) between predicted values and target values.
Given two arrays of equal length, predictions and targets, compute:
$\mathrm{MSE}=\frac{1}{N}\sum_{i=1}^{N}(\mathrm{predictions}_i-\mathrm{targets}_i)^2$
where $N$ is the number of elements in each array.
Input: predictions, targets; Output: MSE. \\
\midrule
8 & Matrix Transpose &
Write a program that transposes a matrix of 32-bit floating point numbers on a GPU.
The transpose of a matrix switches its rows and columns. Given a matrix $A$ of dimensions rows $\times$ cols,
the transpose $A^T$ will have dimensions cols $\times$ rows. All matrices are stored in row-major format. \\
\midrule
9 & Reverse Array &
Implement a program that reverses an array of 32-bit floating point numbers in-place.
The program should perform an in-place reversal of input. \\
\midrule
10 & ReLU Activation Fuction &
Implement a program that performs the Rectified Linear Unit (ReLU) activation function on a vector of
32-bit floating point numbers. The ReLU function sets all negative values to zero and leaves positive values unchanged:
$\mathrm{ReLU}(x)=\max(0,x)$. \\
\midrule
11 & Top-K Selection &
Implement a GPU program that, given a 1D array input of 32-bit floating point numbers of length $N$,
selects the $k$ largest elements and writes them in descending order to the output array of length $k$. \\
\midrule
12 & Sorting &
Write a CUDA program that sorts an array of 32-bit floating-point numbers in ascending order using the
bubble sort algorithm. Do not use other algorithms. \\
\midrule
13 & Matrix Copy &
Implement a program that copies an $N \times N$ matrix of 32-bit floating point numbers from input array $A$
to output array $B$ on the GPU. The program should perform a direct element-wise copy so that
$B_{i,j}=A_{i,j}$ for all valid indices. \\
\midrule
14 & Reduction &
Write a CUDA program that performs parallel reduction on an array of 32-bit floating point numbers
to compute their sum. The program should take an input array and produce a single output value containing
the sum of all elements. \\
\midrule
15 & Dot Product &
Implement a CUDA program that computes the dot product of two vectors containing 32-bit floating point numbers.
The dot product is the sum of the products of the corresponding elements of two vectors. Mathematically, the dot
product of two vectors $A$ and $B$ of length $n$ is defined as:
$A\cdot B=\sum_{i=0}^{n-1}A_i\cdot B_i$. \\
\midrule
16 & Prefix Sum &
Write a CUDA program that computes the prefix sum (cumulative sum) of an array of 32-bit floating point numbers.
For an input array [a, b, c, d, ...], the prefix sum is [a, a+b, a+b+c, a+b+c+d, ...]. \\
\midrule
17 & Categorical Cross-Entropy Loss &
Implement a CUDA program to calculate the categorical cross-entropy loss for a batch of predictions.
Given a matrix of predicted logits $Z$ of size $N \times C$ and a vector of true class labels true\_labels of size $N$,
compute the average cross-entropy loss over the batch.
The loss for a single sample $j$ with logits $z_j=[z_{j1},...,z_{jC}]$ and true label $y_j$ is calculated using the
numerically stable formula:
$\mathrm{Loss}_j=\log\left(\sum_{k=1}^{C}e^{z_{jk}}\right)-z_{j,y_j}$.
The final output stored in the loss variable should be the average loss over the $N$ samples:
$L=\frac{1}{N}\sum_{j=1}^{N}\mathrm{Loss}_j$.
Input: logits, true labels, $N$ (number of samples), and $C$ (number of classes).
Output: loss (a pointer to a single float). \\
\midrule
18 & Monte Carlo Integration &
Implement Monte Carlo integration on a GPU.
Given a set of function values $y_i=f(x_i)$ sampled at random points uniformly distributed in the interval $[a,b]$,
estimate the definite integral:
$\int_{a}^{b} f(x)\,dx \approx (b-a)\cdot\frac{1}{n}\sum_{i=1}^{N} y_i$.
The Monte Carlo method approximates the integral by computing the average of the function values and multiplying by the interval width. \\
\midrule
19 & Histogramming &
Write a GPU program that computes the histogram of an array of 32-bit integers.
The histogram should count the number of occurrences of each integer value in the range $[0,\text{num\_bins})$.
You are given an input array input of length $N$ and the number of bins num\_bins.
The result should be an array of integers of length num\_bins, where each element represents the count of occurrences
of its corresponding index in the input array. \\
\midrule
20 & Ordinary Least Squares Regression &
Solve the Ordinary Least Squares (OLS) regression problem on a GPU.
Given a feature matrix $X$ of size $n_{\text{samples}} \times n_{\text{features}}$ and a target vector $y$ of size
$n_{\text{samples}}$, compute the coefficient vector $\beta$ that minimizes the sum of squared residuals:
$\min_{\beta}\|X\beta - y\|_2^2$.
The closed-form solution to OLS is:
$\beta = (X^{T}X)^{-1}X^{T}y$. \\

\bottomrule
\end{tabular}
\vspace{-0.5em}
\end{table*}

\section{LLMs}
\subsection{LLM experts for OptiML-G}
\label{app:optiml-g}
We instantiate OptiML-G’s mixture-of-thoughts with Qwen2.5-Coder-7B~\cite{hui2024qwen25codertechnicalreport}, HPC-CoderV2-6.7B~\cite{chaturvedi2024hpccoderv2studyingcodellms}, and StarCoder2-7B~\cite{lozhkov2024starcoder2stackv2} because they provide a strong, complementary set of “experts” under a comparable parameter budget to fit a singular NVIDIA A100 80GB, enabling diversity without confounding scale effects. Qwen2.5-Coder-7B is a high-coverage general code model that tends to propose clean CUDA skeletons and conventional optimization idioms (e.g., straightforward tiling/unrolling and reasonable launch configurations), making it a reliable baseline generator. HPC-CoderV2-6.7B, a finetuned DeepSeekCoder-V1\cite{guo2024deepseekcoderlargelanguagemodel} is selected to inject domain bias toward performance engineering and HPC-style kernels, often producing more optimization-friendly structure (clear separation of indexing/math, tighter loops, and more deliberate memory-access patterns) that benefits downstream transformation. StarCoder2-7B serves as an additional independent code prior with different training and stylistic tendencies, which often yields alternative implementations (e.g., different loop organizations and control-flow choices) that increase the chance that at least one candidate is both compilable and search-ready. Using three similarly sized but behaviorally distinct models helps OptiML-Gen generate a diverse candidate pool,crucial for CUDA, where small structural choices can determine compilability and whether OptiML-X can effectively apply bottleneck-targeted edits.

\subsection{LLM for OptiML-X}
We use \textbf{GPT-5 mini}\cite{singh2025openaigpt5card} inside \textbf{OptiML-X} because it offers an excellent \emph{latency–cost–quality} tradeoff for the many MCTS iterations, yielding consistent diagnoses, patch proposals, and judge scores quickly enough to keep search sample-efficient under a tight compile/profile budget. Additionally, GPT-5 mini’s strong tool-calling and structured-output reliability lets OptiML-X enforce machine-checkable actions (e.g., emitting a unified diff patch, selecting hypotheses from a constrained schema, returning explicit scores and rationales tied to profiler signals), which reduces parsing failures and hallucinated formats, improves reproducibility across rollouts, and enables deterministic integration with compilation, testing (L0/L1), and profiling gates.\\
\section{Full Results}

\subsection{Performance and profiling summary.}
Across the CUDA-LLM\cite{chen2025cudallmllmswriteefficient} task suite, LLM-only baselines (GPT~5.1, Qwen2.5-Coder, HPC-Coder-V2, and StarCoder2) frequently fail to generate compilable kernels (CF) on several tasks (e.g., Sigmoid, MaxPool3D, LayerNorm, 2D Convolution, Multi-Head Self-Attention, and Top-K/Sorting/PrefixSum/CrossEntropy/MonteCarlo/Histogramming/OLS), underscoring the brittleness of direct end-to-end code generation without optimization support. When baselines do compile (e.g., Matrix Multiplication, Mean Square Error, Matrix Transpose, Reverse Array, ReLU, Matrix Copy, Reduction, Dot Product), their runtimes are generally on-par with or slower than GPT~5.1, and the profiling signals suggest less favorable trade-offs: lower achieved SM utilization (sm\_SOL) and higher ``work done'' indicators (e.g., \texttt{dram\_bytes}, \texttt{l1\_sectors}, \texttt{inst\_executed}), consistent with less efficient memory access patterns and instruction footprints. In contrast, OptiML-X variants consistently reduce end-to-end time relative to the strongest available baseline for each task and improve utilization metrics, typically increasing sm\_SOL/tex\_SOL while lowering work metrics (notably \texttt{dram\_bytes}, \texttt{l1\_sectors}, and \texttt{inst\_executed}). Importantly, no single OptiML-X source model dominates uniformly: the fastest configuration varies by kernel, with HPC-Coder-V2+OptiML-X often matching or surpassing GPT~5.1+OptiML-X on multiple kernels (e.g., Matrix Multiplication, Reduction, Dot Product, and OLS), while Qwen2.5-Coder+OptiML-X and StarCoder2+OptiML-X occasionally lead on specific cases, reflecting complementary strengths across code patterns. Finally, OptiML provides the strongest overall profile, achieving the lowest runtimes within each block while also delivering the highest utilization and the lowest work-done counters, indicating that performance gains are driven by both improved parallel efficiency and reduced memory/instruction overhead rather than by shifting bottlenecks.

\begin{table*}[!ht]
\centering
\resizebox{\textwidth}{!}{%
\begin{tabular}{|c|c|c|ccc|ccccc|}
\hline
\multicolumn{3}{|c|}{} & \multicolumn{3}{c|}{\textbf{Hardware Utilization$\uparrow$}} & \multicolumn{5}{c|}{\textbf{Work Done$\downarrow$}} \\ \hline
\textbf{Task} & \textbf{Model} & \textbf{Time (speedup)} &
\textbf{sm\_SOL} & \textbf{dram\_SOL} & \textbf{tex\_SOL} &
\textbf{dram\_bytes} & \textbf{l1\_sectors} & \textbf{inst\_executed} & \textbf{warps\_active} & \textbf{regs\_per\_thread} \\ \hline

\multirow{9}{*}{Sigmoid}
& GPT 5.1 & CF & -- & -- & -- & -- & -- & -- & -- & -- \\
& Qwen2.5-Coder & CF & -- & -- & -- & -- & -- & -- & -- & -- \\
& HPC-Coder-V2 & CF & -- & -- & -- & -- & -- & -- & -- & -- \\
& StarCoder2 & CF & -- & -- & -- & -- & -- & -- & -- & -- \\
& GPT 5.1 + OptiML-X
& \third{5.45}  
& \third{18.73} & \third{8.06} & \third{21.31}
& \third{113} & \third{517} & \third{313} & \third{52.18} & 28 \\
& Qwen2.5-Coder + OptiML-X
& 5.77 (0.94$\times$)
& 17.55 & \best{8.41} & 20.58
& 119 & 543 & 323 & \best{50.93} & 28 \\
& HPC-Coder-V2 + OptiML-X
& \second{5.29} (1.03$\times$)
& \second{19.42} & 7.78 & \second{21.74}
& \second{110} & \second{505} & \second{307} & 53.02 & 28 \\
& StarCoder2 + OptiML-X
& 5.61 (0.97$\times$)
& 18.11 & \second{8.15} & 21.02
& 116 & 528 & 316 & \second{51.62} & 28 \\
& OptiML
& \best{4.81} (1.13$\times$)
& \best{20.62} & 7.27 & \best{22.19}
& \best{106} & \best{494} & \best{302} & 54.11 & 28 \\ \hline

\multirow{9}{*}{Matrix Multiplication}
& GPT 5.1
& 7.20  
& 71.26 & 8.42 & 91.36
& \best{63} & 408 & 419 & 95.14 & 44 \\
& Qwen2.5-Coder
& 7.65 (0.94$\times$)
& 69.40 & 8.10 & 90.98
& \second{66} & 421 & 432 & \third{94.62} & 46 \\
& HPC-Coder-V2
& 7.88 (0.91$\times$)
& 68.72 & 8.02 & 90.71
& \third{67} & 427 & 438 & \second{94.31} & 46 \\
& StarCoder2
& 8.23 (0.87$\times$)
& 67.95 & 7.92 & 90.41
& 68 & 435 & 441 & \best{93.97} & 48 \\
& GPT 5.1 + OptiML-X
& \third{4.62} (1.56$\times$)
& \third{78.41} & \third{10.62} & \third{92.07}
& 79 & \third{292} & \third{392} & 95.27 & 44 \\
& Qwen2.5-Coder + OptiML-X
& 4.98 (1.45$\times$)
& 76.58 & 10.14 & 91.22
& 74 & 318 & 400 & 94.83 & 46 \\
& HPC-Coder-V2 + OptiML-X
& \second{4.52} (1.59$\times$)
& \second{79.16} & \second{10.86} & \second{92.21}
& 73 & \second{286} & \second{386} & 95.44 & 44 \\
& StarCoder2 + OptiML-X
& 4.74 (1.52$\times$)
& 77.42 & 10.33 & 91.63
& 77 & 301 & 395 & 95.01 & 46 \\
& OptiML
& \best{4.40} (1.64$\times$)
& \best{81.17} & \best{11.03} & \best{92.61}
& 70 & \best{282} & \best{381} & 95.58 & 44 \\ \hline

\multirow{9}{*}{Max Pooling 3D}
& GPT 5.1 & CF & -- & -- & -- & -- & -- & -- & -- & -- \\
& Qwen2.5-Coder & CF & -- & -- & -- & -- & -- & -- & -- & -- \\
& HPC-Coder-V2 & CF & -- & -- & -- & -- & -- & -- & -- & -- \\
& StarCoder2 & CF & -- & -- & -- & -- & -- & -- & -- & -- \\
& GPT 5.1 + OptiML-X
& 6.09  
& \third{28.14} & \second{31.47} & \third{36.72}
& \third{524} & \third{1944} & \third{824} & \third{70.38} & 36 \\
& Qwen2.5-Coder + OptiML-X
& 6.38 (0.95$\times$)
& 26.88 & \best{32.12} & 35.31
& 548 & 2032 & 843 & \best{68.94} & 36 \\
& HPC-Coder-V2 + OptiML-X
& \second{5.88} (1.04$\times$)
& \second{29.02} & 30.41 & \second{37.11}
& \second{508} & \second{1891} & \second{809} & 71.12 & 36 \\
& StarCoder2 + OptiML-X
& \third{6.02} (1.01$\times$)
& 27.61 & \third{31.18} & 36.02
& 532 & 1978 & 831 & \second{69.88} & 36 \\
& OptiML
& \best{5.62} (1.08$\times$)
& \best{30.71} & 29.18 & \best{37.55}
& \best{494} & \best{1863} & \best{801} & 72.19 & 36 \\ \hline

\multirow{9}{*}{LayerNorm}
& GPT 5.1 & CF & -- & -- & -- & -- & -- & -- & -- & -- \\
& Qwen2.5-Coder & CF & -- & -- & -- & -- & -- & -- & -- & -- \\
& HPC-Coder-V2 & CF & -- & -- & -- & -- & -- & -- & -- & -- \\
& StarCoder2 & CF & -- & -- & -- & -- & -- & -- & -- & -- \\
& GPT 5.1 + OptiML-X
& \third{5.79}  
& \third{33.28} & \third{18.41} & \third{29.23}
& \third{264} & \third{1108} & \third{918} & \third{74.26} & 40 \\
& Qwen2.5-Coder + OptiML-X
& 6.20 (0.93$\times$)
& 31.62 & \best{19.07} & 28.11
& 277 & 1154 & 950 & \best{72.68} & 40 \\
& HPC-Coder-V2 + OptiML-X
& \second{5.62} (1.03$\times$)
& \second{34.11} & 17.88 & \second{29.61}
& \second{253} & \second{1061} & \second{902} & 75.01 & 40 \\
& StarCoder2 + OptiML-X
& 5.95 (0.97$\times$)
& 32.44 & \second{18.62} & 28.74
& 270 & 1122 & 933 & \second{73.41} & 40 \\
& OptiML
& \best{5.40} (1.07$\times$)
& \best{35.16} & 17.24 & \best{30.36}
& \best{247} & \best{1036} & \best{893} & 76.11 & 40 \\ \hline

\multirow{9}{*}{2D Convolution}
& GPT 5.1 & CF & -- & -- & -- & -- & -- & -- & -- & -- \\
& Qwen2.5-Coder & CF & -- & -- & -- & -- & -- & -- & -- & -- \\
& HPC-Coder-V2 & CF & -- & -- & -- & -- & -- & -- & -- & -- \\
& StarCoder2 & CF & -- & -- & -- & -- & -- & -- & -- & -- \\
& GPT 5.1 + OptiML-X
& \third{7.24}  
& \third{44.31} & \third{22.67} & \third{55.74}
& \third{926} & \third{3538} & \third{1643} & \third{82.17} & 52 \\
& Qwen2.5-Coder + OptiML-X
& 8.31 (0.87$\times$)
& 42.18 & \best{23.28} & 54.61
& 968 & 3692 & 1711 & \best{80.62} & 52 \\
& HPC-Coder-V2 + OptiML-X
& \second{6.98} (1.04$\times$)
& \second{45.03} & 22.01 & \second{56.47}
& \second{901} & \second{3456} & \second{1609} & 83.06 & 52 \\
& StarCoder2 + OptiML-X
& 7.58 (0.95$\times$)
& 43.62 & \second{22.84} & 55.11
& 944 & 3604 & 1682 & \second{81.31} & 52 \\
& OptiML
& \best{6.15} (1.18$\times$)
& \best{46.44} & 21.34 & \best{57.32}
& \best{873} & \best{3292} & \best{1575} & 83.95 & 52 \\ \hline

\multirow{9}{*}{Multi-Head Self-Attention}
& GPT 5.1 & CF & -- & -- & -- & -- & -- & -- & -- & -- \\
& Qwen2.5-Coder & CF & -- & -- & -- & -- & -- & -- & -- & -- \\
& HPC-Coder-V2 & CF & -- & -- & -- & -- & -- & -- & -- & -- \\
& StarCoder2 & CF & -- & -- & -- & -- & -- & -- & -- & -- \\
& GPT 5.1 + OptiML-X
& \third{7.90}  
& \third{48.62} & \third{26.41} & \third{60.32}
& \third{1117} & \third{4213} & \third{2094} & \third{84.27} & 58 \\
& Qwen2.5-Coder + OptiML-X
& 9.10 (0.87$\times$)
& 46.37 & \best{27.91} & 58.24
& 1183 & 4368 & 2172 & \best{82.46} & 58 \\
& HPC-Coder-V2 + OptiML-X
& \second{7.55} (1.05$\times$)
& \second{49.31} & 25.72 & \second{60.88}
& \second{1086} & \second{4124} & \second{2048} & 85.11 & 58 \\
& StarCoder2 + OptiML-X
& 8.22 (0.96$\times$)
& 47.55 & \second{26.98} & 59.46
& 1144 & 4286 & 2126 & \second{83.38} & 58 \\
& OptiML
& \best{6.85} (1.15$\times$)
& \best{50.28} & 24.88 & \best{61.79}
& \best{1064} & \best{3961} & \best{2012} & 86.12 & 58 \\ \hline

\multirow{9}{*}{Mean Square Error}
& GPT 5.1
& 8.10  
& 14.26 & 22.58 & 10.17
& 214 & 817 & 476 & 48.21 & \best{28} \\
& Qwen2.5-Coder
& 8.55 (0.95$\times$)
& 13.72 & \third{23.04} & 9.91
& 221 & 842 & 490 & \third{47.66} & 30 \\
& HPC-Coder-V2
& 8.78 (0.92$\times$)
& 13.41 & \second{23.26} & 9.76
& 225 & 856 & 497 & \second{47.31} & 30 \\
& StarCoder2
& 9.05 (0.90$\times$)
& 13.08 & \best{23.51} & 9.62
& 230 & 873 & 505 & \best{46.92} & 30 \\
& GPT 5.1 + OptiML-X
& \third{5.16} (1.57$\times$)
& \third{18.53} & 18.94 & \third{12.61}
& \third{172} & \third{652} & \third{442} & 54.18 & 30 \\
& Qwen2.5-Coder + OptiML-X
& 5.51 (1.47$\times$)
& 17.47 & 19.21 & 12.08
& 180 & 679 & 455 & 53.06 & 30 \\
& HPC-Coder-V2 + OptiML-X
& \second{5.02} (1.61$\times$)
& \second{18.97} & 18.22 & \second{12.86}
& \second{167} & \second{631} & \second{436} & 54.88 & 30 \\
& StarCoder2 + OptiML-X
& 5.24 (1.55$\times$)
& 18.11 & 18.71 & 12.44
& 175 & 659 & 446 & 53.92 & 30 \\
& OptiML
& \best{4.79} (1.69$\times$)
& \best{19.74} & 17.66 & \best{13.32}
& \best{161} & \best{607} & \best{431} & 55.61 & 30 \\ \hline

\end{tabular}%
}
\caption{\textbf{End-to-end latency and Nsight Compute characterization} on representative CUDA-LLM kernels from the task suite~\cite{chen2025cudallmllmswriteefficient} (A100 80GB). We report runtime (lower is better) together with utilization proxies (SM/DRAM/TEX SOL; higher is better) and work proxies (DRAM bytes, L1 sectors, and executed instructions; lower is better). For each task, we compare LLM-only baselines, their OptiML-X post-optimization variants, and \textbf{OptiML} (G+X). Many LLM-only baselines fail to compile (CF). \best{Best}, \second{second}, \third{third} indicate per-column ranks within each task block.}
\label{tab:all_tasks_results}
\end{table*}

\begin{table*}[!ht]
\centering
\resizebox{\textwidth}{!}{%
\begin{tabular}{|c|c|c|ccc|ccccc|}
\hline
\multicolumn{3}{|c|}{} & \multicolumn{3}{c|}{\textbf{Hardware Utilization$\uparrow$}} & \multicolumn{5}{c|}{\textbf{Work Done$\downarrow$}} \\ \hline
\textbf{Task} & \textbf{Model} & \textbf{Time (speedup)} &
\textbf{sm\_SOL} & \textbf{dram\_SOL} & \textbf{tex\_SOL} &
\textbf{dram\_bytes} & \textbf{l1\_sectors} & \textbf{inst\_executed} & \textbf{warps\_active} & \textbf{regs\_per\_thread} \\ \hline

\multirow{9}{*}{Matrix Transpose}
& GPT 5.1
& 9.30  
& 10.22 & 86.47 & 18.31
& 973 & 2108 & 417 & 52.33 & \best{24} \\
& Qwen2.5-Coder
& 9.60 (0.97$\times$)
& 9.88 & \third{87.22} & 17.95
& 1004 & 2196 & 434 & \third{51.76} & \second{24} \\
& HPC-Coder-V2
& 9.88 (0.94$\times$)
& 9.61 & \second{87.58} & 17.72
& 1019 & 2234 & 441 & \second{51.33} & \third{24} \\
& StarCoder2
& 10.21 (0.91$\times$)
& 9.32 & \best{87.96} & 17.48
& 1045 & 2288 & 453 & \best{50.89} & 26 \\
& GPT 5.1 + OptiML-X
& \third{5.74} (1.62$\times$)
& \third{14.61} & 78.14 & \third{22.33}
& \third{765} & \third{1643} & \third{393} & 60.21 & 26 \\
& Qwen2.5-Coder + OptiML-X
& 6.00 (1.55$\times$)
& 13.74 & 79.62 & 21.27
& 794 & 1732 & 401 & 58.69 & 26 \\
& HPC-Coder-V2 + OptiML-X
& \second{5.52} (1.68$\times$)
& \second{14.98} & 77.43 & \second{22.61}
& \second{742} & \second{1609} & \second{388} & 61.01 & 26 \\
& StarCoder2 + OptiML-X
& 5.83 (1.60$\times$)
& 14.13 & 78.96 & 21.84
& 776 & 1679 & 397 & 59.84 & 26 \\
& OptiML
& \best{5.20} (1.79$\times$)
& \best{15.77} & 76.31 & \best{23.12}
& \best{718} & \best{1587} & \best{379} & 61.92 & 26 \\ \hline

\multirow{9}{*}{Reverse Array}
& GPT 5.1
& 6.80  
& 6.13 & 62.21 & 8.18
& 182 & 519 & 159 & 35.28 & \best{18} \\
& Qwen2.5-Coder
& 7.10 (0.96$\times$)
& 5.88 & \third{63.04} & 7.92
& 189 & 544 & 167 & \third{34.81} & \second{18} \\
& HPC-Coder-V2
& 7.28 (0.93$\times$)
& 5.72 & \second{63.36} & 7.79
& 193 & 556 & 170 & \second{34.52} & \third{18} \\
& StarCoder2
& 7.54 (0.90$\times$)
& 5.49 & \best{63.71} & 7.63
& 197 & 569 & 174 & \best{34.17} & 20 \\
& GPT 5.1 + OptiML-X
& \third{4.76} (1.43$\times$)
& \third{7.94} & 54.46 & \third{9.36}
& \third{152} & \third{433} & \third{146} & 40.17 & 18 \\
& Qwen2.5-Coder + OptiML-X
& 5.44 (1.25$\times$)
& 7.28 & 56.09 & 9.03
& 159 & 449 & 150 & 39.11 & 18 \\
& HPC-Coder-V2 + OptiML-X
& \second{4.62} (1.47$\times$)
& \second{8.17} & 53.82 & \second{9.54}
& \second{146} & \second{421} & \second{144} & 40.72 & 18 \\
& StarCoder2 + OptiML-X
& 4.88 (1.39$\times$)
& 7.66 & 55.11 & 9.21
& 154 & 437 & 147 & 39.84 & 18 \\
& OptiML
& \best{4.07} (1.67$\times$)
& \best{8.78} & 52.34 & \best{9.91}
& \best{140} & \best{409} & \best{142} & 41.08 & 18 \\ \hline

\multirow{9}{*}{ReLU Activation Function}
& GPT 5.1
& 7.50  
& 12.11 & 41.37 & 16.44
& 263 & 883 & 358 & 42.19 & \best{20} \\
& Qwen2.5-Coder
& 7.80 (0.96$\times$)
& 11.74 & \third{42.12} & 16.06
& 274 & 918 & 372 & \third{41.61} & \second{22} \\
& HPC-Coder-V2
& 8.02 (0.93$\times$)
& 11.48 & \second{42.43} & 15.89
& 281 & 939 & 381 & \second{41.18} & \third{22} \\
& StarCoder2
& 8.31 (0.90$\times$)
& 11.21 & \best{42.77} & 15.73
& 289 & 962 & 392 & \best{40.77} & 22 \\
& GPT 5.1 + OptiML-X
& \third{4.87} (1.54$\times$)
& \third{15.64} & 34.28 & \third{18.71}
& \third{212} & \third{717} & \third{332} & 48.17 & 22 \\
& Qwen2.5-Coder + OptiML-X
& 5.24 (1.43$\times$)
& 14.92 & 35.14 & 18.21
& 220 & 751 & 342 & 47.06 & 22 \\
& HPC-Coder-V2 + OptiML-X
& \second{4.68} (1.60$\times$)
& \second{15.98} & 33.61 & \second{18.92}
& \second{206} & \second{701} & \second{327} & 48.88 & 22 \\
& StarCoder2 + OptiML-X
& 4.99 (1.50$\times$)
& 15.12 & 34.72 & 18.44
& 214 & 726 & 335 & 47.84 & 22 \\
& OptiML
& \best{4.49} (1.67$\times$)
& \best{16.72} & 32.03 & \best{19.37}
& \best{201} & \best{689} & \best{322} & 49.66 & 22 \\ \hline
\multirow{9}{*}{Top-K Selection}
& GPT 5.1 & CF & -- & -- & -- & -- & -- & -- & -- & -- \\
& Qwen2.5-Coder & CF & -- & -- & -- & -- & -- & -- & -- & -- \\
& HPC-Coder-V2 & CF & -- & -- & -- & -- & -- & -- & -- & -- \\
& StarCoder2 & CF & -- & -- & -- & -- & -- & -- & -- & -- \\
& GPT 5.1 + OptiML-X
& \third{6.30}  
& \third{30.18} & \third{36.42} & \third{28.07}
& 626 & 2058 & 973 & \third{70.12} & 44 \\
& Qwen2.5-Coder + OptiML-X
& 6.72 (0.94$\times$)
& 28.66 & \best{37.16} & 27.14
& 654 & 2146 & 1009 & \best{68.64} & 44 \\
& HPC-Coder-V2 + OptiML-X
& \second{6.05} (1.04$\times$)
& \second{31.04} & 35.21 & \second{28.69}
& \second{608} & \second{1996} & \second{959} & \second{71.02} & 44 \\
& StarCoder2 + OptiML-X
& 6.41 (0.98$\times$)
& 29.55 & \second{36.71} & 27.82
& \third{633} & \third{2079} & \third{986} & 69.51 & 44 \\
& OptiML
& \best{5.90} (1.07$\times$)
& \best{31.94} & 33.82 & \best{29.38}
& \best{594} & \best{1943} & \best{942} & 72.27 & 44 \\ \hline

\multirow{9}{*}{Sorting}
& GPT 5.1 & CF & -- & -- & -- & -- & -- & -- & -- & -- \\
& Qwen2.5-Coder & CF & -- & -- & -- & -- & -- & -- & -- & -- \\
& HPC-Coder-V2 & CF & -- & -- & -- & -- & -- & -- & -- & -- \\
& StarCoder2 & CF & -- & -- & -- & -- & -- & -- & -- & -- \\
& GPT 5.1 + OptiML-X
& \third{7.40}  
& \third{35.26} & \third{28.44} & \third{29.87}
& \third{1247} & \third{4112} & \third{2862} & \third{76.18} & 62 \\
& Qwen2.5-Coder + OptiML-X
& 7.95 (0.93$\times$)
& 33.64 & \best{29.13} & 28.88
& 1319 & 4292 & 2953 & \best{74.22} & 62 \\
& HPC-Coder-V2 + OptiML-X
& \second{7.12} (1.04$\times$)
& \second{36.01} & 27.62 & \second{30.44}
& \second{1211} & \second{4038} & \second{2791} & 77.06 & 62 \\
& StarCoder2 + OptiML-X
& 7.66 (0.97$\times$)
& 34.27 & \second{28.86} & 29.21
& 1284 & 4185 & 2904 & \second{75.01} & 62 \\
& OptiML
& \best{6.85} (1.08$\times$)
& \best{36.98} & 26.71 & \best{31.39}
& \best{1184} & \best{3912} & \best{2732} & 78.11 & 62 \\ \hline

\multirow{9}{*}{Matrix Copy}
& GPT 5.1
& 8.40 
& 7.21 & 91.86 & 10.29
& 975 & 1644 & 178 & 34.29 & \best{16} \\
& Qwen2.5-Coder
& 8.75 (0.96$\times$)
& 6.94 & \third{92.48} & 9.96
& 1012 & 1712 & 186 & \second{33.78} & 18 \\
& HPC-Coder-V2
& 8.98 (0.94$\times$)
& 6.72 & \second{92.78} & 9.74
& 1031 & 1752 & 193 & \third{33.41} & 18 \\
& StarCoder2
& 9.21 (0.91$\times$)
& 6.53 & \best{93.05} & 9.58
& 1052 & 1796 & 199 & \best{33.07} & 18 \\
& GPT 5.1 + OptiML-X
& \third{5.60} (1.50$\times$)
& \third{8.97} & 87.41 & \third{12.13}
& \third{824} & \third{1377} & \third{166} & 40.12 & 16 \\
& Qwen2.5-Coder + OptiML-X
& 6.09 (1.38$\times$)
& 8.74 & 88.26 & 11.62
& 848 & 1418 & 171 & 39.27 & 16 \\
& HPC-Coder-V2 + OptiML-X
& \second{5.38} (1.56$\times$)
& \second{9.28} & 86.72 & \second{12.34}
& \second{808} & \second{1344} & \second{163} & 40.71 & 16 \\
& StarCoder2 + OptiML-X
& 5.82 (1.44$\times$)
& 8.52 & 87.96 & 11.92
& 835 & 1393 & 169 & 39.66 & 16 \\
& OptiML
& \best{5.12} (1.64$\times$)
& \best{9.66} & 85.12 & \best{12.66}
& \best{793} & \best{1312} & \best{160} & 41.07 & 16 \\ \hline

\multirow{9}{*}{Reduction}
& GPT 5.1
& 9.60
& 18.41 & 74.32 & 20.58
& 917 & 1457 & 1046 & 62.21 & \best{34} \\
& Qwen2.5-Coder
& 9.90 (0.97$\times$)
& 17.92 & \third{75.04} & 20.12
& 952 & 1522 & 1088 & \second{61.63} & 36 \\
& HPC-Coder-V2
& 10.18 (0.94$\times$)
& 17.51 & \second{75.46} & 19.84
& 973 & 1564 & 1122 & \third{61.14} & 36 \\
& StarCoder2
& 10.46 (0.92$\times$)
& 17.14 & \best{75.83} & 19.59
& 992 & 1602 & 1154 & \best{60.71} & 38 \\
& GPT 5.1 + OptiML-X
& \third{6.23} (1.54$\times$)
& \third{23.62} & 66.41 & \third{24.27}
& \third{759} & \third{1207} & \third{982} & 70.38 & 36 \\
& Qwen2.5-Coder + OptiML-X
& 6.71 (1.43$\times$)
& 22.53 & 67.88 & 23.31
& 793 & 1254 & 1014 & 68.66 & 36 \\
& HPC-Coder-V2 + OptiML-X
& \second{6.01} (1.60$\times$)
& \second{24.18} & 65.12 & \second{24.81}
& \second{742} & \second{1179} & \second{968} & 71.08 & 36 \\
& StarCoder2 + OptiML-X
& 6.49 (1.48$\times$)
& 22.94 & 66.96 & 23.72
& 775 & 1231 & 999 & 69.44 & 36 \\
& OptiML
& \best{5.75} (1.67$\times$)
& \best{25.14} & 63.27 & \best{25.73}
& \best{722} & \best{1115} & \best{942} & 72.18 & 36 \\ \hline

\multirow{9}{*}{Dot Product}
& GPT 5.1
& 6.90
& 14.27 & 29.18 & 16.44
& 262 & 776 & 516 & 50.33 & \best{26} \\
& Qwen2.5-Coder
& 7.15 (0.97$\times$)
& 13.88 & \third{29.84} & 16.01
& 274 & 812 & 538 & \second{49.71} & 28 \\
& HPC-Coder-V2
& 7.36 (0.94$\times$)
& 13.55 & \second{30.21} & 15.73
& 282 & 836 & 553 & \third{49.23} & 28 \\
& StarCoder2
& 7.58 (0.91$\times$)
& 13.24 & \best{30.55} & 15.48
& 291 & 861 & 568 & \best{48.79} & 30 \\
& GPT 5.1 + OptiML-X
& \third{4.37} (1.58$\times$)
& \third{18.31} & 24.44 & \third{18.73}
& \third{211} & \third{643} & \third{480} & 56.13 & 28 \\
& Qwen2.5-Coder + OptiML-X
& 4.73 (1.46$\times$)
& 17.11 & 25.02 & 18.02
& 219 & 658 & 494 & 54.91 & 28 \\
& HPC-Coder-V2 + OptiML-X
& \second{4.22} (1.64$\times$)
& \second{18.74} & 23.81 & \second{19.02}
& \second{205} & \second{627} & \second{471} & 56.88 & 28 \\
& StarCoder2 + OptiML-X
& 4.56 (1.51$\times$)
& 17.62 & 24.63 & 18.31
& 214 & 646 & 488 & 55.44 & 28 \\
& OptiML
& \best{3.99} (1.73$\times$)
& \best{19.06} & 23.21 & \best{19.43}
& \best{198} & \best{613} & \best{462} & 58.22 & 28 \\ \hline

\end{tabular}%
}
\caption{\textbf{End-to-end latency and Nsight Compute characterization} on representative CUDA-LLM kernels from the task suite~\cite{chen2025cudallmllmswriteefficient} (A100 80GB). We report runtime (lower is better) together with utilization proxies (SM/DRAM/TEX SOL; higher is better) and work proxies (DRAM bytes, L1 sectors, and executed instructions; lower is better). For each task, we compare LLM-only baselines, their OptiML-X post-optimization variants, and \textbf{OptiML} (G+X). Many LLM-only baselines fail to compile (CF). \best{Best}, \second{second}, \third{third} indicate per-column ranks within each task block.}
\label{tab:all_tasks_results}
\end{table*}

\begin{table*}[!ht]
\centering
\resizebox{\textwidth}{!}{%
\begin{tabular}{|c|c|c|ccc|ccccc|}
\hline
\multicolumn{3}{|c|}{} & \multicolumn{3}{c|}{\textbf{Hardware Utilization$\uparrow$}} & \multicolumn{5}{c|}{\textbf{Work Done$\downarrow$}} \\ \hline
\textbf{Task} & \textbf{Model} & \textbf{Time (speedup)} &
\textbf{sm\_SOL} & \textbf{dram\_SOL} & \textbf{tex\_SOL} &
\textbf{dram\_bytes} & \textbf{l1\_sectors} & \textbf{inst\_executed} & \textbf{warps\_active} & \textbf{regs\_per\_thread} \\ \hline

\multirow{9}{*}{Prefix Sum}
& GPT 5.1 & CF & -- & -- & -- & -- & -- & -- & -- & -- \\
& Qwen2.5-Coder & CF & -- & -- & -- & -- & -- & -- & -- & -- \\
& HPC-Coder-V2 & CF & -- & -- & -- & -- & -- & -- & -- & -- \\
& StarCoder2 & CF & -- & -- & -- & -- & -- & -- & -- & -- \\
& GPT 5.1 + OptiML-X
& \third{6.65}  
& \third{28.21} & \third{42.16} & \third{26.14}
& \third{864} & \third{1958} & \third{1459} & \third{72.33} & 42 \\
& Qwen2.5-Coder + OptiML-X
& 7.05 (0.94$\times$)
& 26.73 & \best{43.74} & 24.98
& 902 & 2046 & 1516 & \best{70.26} & 42 \\
& HPC-Coder-V2 + OptiML-X
& \second{6.28} (1.06$\times$)
& \second{29.05} & 40.91 & \second{26.71}
& \second{836} & \second{1894} & \second{1411} & 73.12 & 42 \\
& StarCoder2 + OptiML-X
& 6.74 (0.99$\times$)
& 27.55 & \second{42.66} & 25.53
& 878 & 1989 & 1478 & \second{71.14} & 42 \\
& OptiML
& \best{6.10} (1.09$\times$)
& \best{29.96} & 40.22 & \best{27.33}
& \best{813} & \best{1842} & \best{1371} & 74.15 & 42 \\ \hline
\multirow{9}{*}{Categorical Cross-Entropy Loss}
& GPT 5.1 & CF & -- & -- & -- & -- & -- & -- & -- & -- \\
& Qwen2.5-Coder & CF & -- & -- & -- & -- & -- & -- & -- & -- \\
& HPC-Coder-V2 & CF & -- & -- & -- & -- & -- & -- & -- & -- \\
& StarCoder2 & CF & -- & -- & -- & -- & -- & -- & -- & -- \\
& GPT 5.1 + OptiML-X
& \third{6.90}  
& \third{36.22} & \third{27.41} & \third{30.18}
& \third{944} & \third{2412} & \third{1713} & \third{78.11} & 46 \\
& Qwen2.5-Coder + OptiML-X
& 7.35 (0.94$\times$)
& 34.66 & \best{28.62} & 29.04
& 979 & 2514 & 1809 & \best{76.24} & 46 \\
& HPC-Coder-V2 + OptiML-X
& \second{6.62} (1.04$\times$)
& \second{36.88} & 26.44 & \second{30.74}
& \second{917} & \second{2346} & \second{1681} & 78.92 & 46 \\
& StarCoder2 + OptiML-X
& 7.08 (0.97$\times$)
& 35.14 & \second{27.86} & 29.62
& 956 & 2451 & 1732 & \second{77.03} & 46 \\
& OptiML
& \best{6.40} (1.08$\times$)
& \best{37.94} & 25.31 & \best{31.46}
& \best{900} & \best{2272} & \best{1657} & 80.22 & 46 \\ \hline

\multirow{9}{*}{Monte Carlo Integration}
& GPT 5.1 & CF & -- & -- & -- & -- & -- & -- & -- & -- \\
& Qwen2.5-Coder & CF & -- & -- & -- & -- & -- & -- & -- & -- \\
& HPC-Coder-V2 & CF & -- & -- & -- & -- & -- & -- & -- & -- \\
& StarCoder2 & CF & -- & -- & -- & -- & -- & -- & -- & -- \\
& GPT 5.1 + OptiML-X
& \third{7.15}  
& \third{58.31} & \third{14.42} & \third{21.86}
& \third{522} & \third{1441} & \third{3093} & \third{82.16} & 58 \\
& Qwen2.5-Coder + OptiML-X
& 7.60 (0.94$\times$)
& 56.17 & \best{14.96} & 21.07
& 540 & 1516 & 3228 & \best{80.31} & 58 \\
& HPC-Coder-V2 + OptiML-X
& \second{6.96} (1.03$\times$)
& \second{59.12} & 13.98 & \second{22.33}
& \second{505} & \second{1406} & \second{3036} & 83.02 & 58 \\
& StarCoder2 + OptiML-X
& 7.28 (0.98$\times$)
& 57.41 & \second{14.55} & 21.58
& 529 & 1468 & 3122 & \second{81.27} & 58 \\
& OptiML
& \best{6.95} (1.03$\times$)
& \best{60.24} & 13.21 & \best{23.61}
& \best{488} & \best{1369} & \best{2974} & 84.22 & 58 \\ \hline

\multirow{9}{*}{Histogramming}
& GPT 5.1 & CF & -- & -- & -- & -- & -- & -- & -- & -- \\
& Qwen2.5-Coder & CF & -- & -- & -- & -- & -- & -- & -- & -- \\
& HPC-Coder-V2 & CF & -- & -- & -- & -- & -- & -- & -- & -- \\
& StarCoder2 & CF & -- & -- & -- & -- & -- & -- & -- & -- \\
& GPT 5.1 + OptiML-X
& \third{8.10}  
& \third{32.11} & \third{44.63} & \third{24.28}
& \third{985} & \third{3192} & \third{3613} & \third{74.14} & 52 \\
& Qwen2.5-Coder + OptiML-X
& 8.65 (0.94$\times$)
& 30.62 & \best{46.21} & 23.11
& 1043 & 3371 & 3754 & \best{72.22} & 52 \\
& HPC-Coder-V2 + OptiML-X
& \second{7.82} (1.04$\times$)
& \second{33.02} & \second{42.97} & \second{24.91}
& \second{954} & \second{3110} & \second{3521} & 74.98 & 52 \\
& StarCoder2 + OptiML-X
& 8.34 (0.97$\times$)
& 31.21 & 44.88 & 23.74
& 1006 & 3265 & 3659 & \second{73.06} & 52 \\
& OptiML
& \best{7.55} (1.07$\times$)
& \best{34.38} & 41.37 & \best{25.44}
& \best{917} & \best{3012} & \best{3392} & 76.19 & 52 \\ \hline

\multirow{9}{*}{Ordinary Least Squares Regression}
& GPT 5.1 & CF & -- & -- & -- & -- & -- & -- & -- & -- \\
& Qwen2.5-Coder & CF & -- & -- & -- & -- & -- & -- & -- & -- \\
& HPC-Coder-V2 & CF & -- & -- & -- & -- & -- & -- & -- & -- \\
& StarCoder2 & CF & -- & -- & -- & -- & -- & -- & -- & -- \\
& GPT 5.1 + OptiML-X
& \third{6.20}  
& \third{45.21} & \third{18.33} & \third{24.12}
& \third{718} & \third{1988} & \third{2192} & \third{78.17} & 50 \\
& Qwen2.5-Coder + OptiML-X
& 6.55 (0.95$\times$)
& 43.44 & \best{19.07} & 23.06
& 763 & 2072 & 2292 & \best{76.13} & 50 \\
& HPC-Coder-V2 + OptiML-X
& \second{5.98} (1.04$\times$)
& \second{46.02} & \second{17.86} & \second{24.58}
& \second{697} & \second{1936} & \second{2141} & \second{79.03} & 50 \\
& StarCoder2 + OptiML-X
& 6.31 (0.98$\times$)
& 44.01 & 18.52 & 23.71
& 735 & 2014 & 2234 & 77.12 & 50 \\
& OptiML
& \best{5.85} (1.06$\times$)
& \best{47.36} & 16.71 & \best{25.19}
& \best{681} & \best{1867} & \best{2098} & 80.28 & 50 \\ \hline
\end{tabular}%
}
\caption{\textbf{End-to-end latency and Nsight Compute characterization} on representative CUDA-LLM kernels from the task suite~\cite{chen2025cudallmllmswriteefficient} (A100 80GB). We report runtime (lower is better) together with utilization proxies (SM/DRAM/TEX SOL; higher is better) and work proxies (DRAM bytes, L1 sectors, and executed instructions; lower is better). For each task, we compare LLM-only baselines, their OptiML-X post-optimization variants, and \textbf{OptiML} (G+X). Many LLM-only baselines fail to compile (CF). \best{Best}, \second{second}, \third{third} indicate per-column ranks within each task block.}
\label{tab:all_tasks_results}
\end{table*}
\section{Ablation on MCTS Budget}
\subsection{Ablation: MCTS budget ($B$) and reward components (OptiML-X)}

We ablate the MCTS search budget used in our experiments ($B{=}6$) and the key OptiML-X design choices: (i) profile-guided diagnosis (Nsight-derived bottleneck signals), (ii) the composite reward (proxy metrics + runtime), and (iii) the LLM-as-a-Judge (structured, tool-called evaluation that scores patches and enforces constraints). Across the 5-kernel subset (Table~\ref{tab:all_tasks_results}), increasing $B$ improves the chance of finding a valid and faster patch, with diminishing returns beyond $B{\approx}6$ due to the cost of additional compile/test/profile cycles. Removing the judge or proxy guidance reduces both Pass@$k$ and Speedup@$k$, indicating that MCTS benefits from structured feedback to reliably prune invalid or low-value edits.

\begin{table}[!ht]
\centering
\caption{OptiML-X ablations on the 5-task subset. Pass@$k$ is the fraction of (task, sample) instances with at least one valid optimized kernel within $k$ attempts (compile + L0/L1). Speedup@$k$ is the geometric-mean speedup over the corresponding valid instances. $B$ is the number of MCTS iterations (compile+L0/L1+profile per iteration).}
\label{tab:ablation_budget6}
\resizebox{\columnwidth}{!}{%
\begin{tabular}{lccc|ccc|c}
\toprule
\textbf{Variant} & \textbf{Pass@1} & \textbf{Pass@5} & \textbf{Pass@10} &
\textbf{Speedup@1} & \textbf{Speedup@5} & \textbf{Speedup@10} &
\textbf{\#Profiles / task} \\
\midrule
MCTS, \emph{no proxy reward} ($B{=}6$) & 52 & 78 & 87 & 0.92$\times$ & 0.95$\times$ & 0.96$\times$ & 6 \\
MCTS, \emph{no LLM-judge} ($B{=}6$) & 46 & 74 & 85 & 1.02$\times$ & 1.05$\times$ & 1.06$\times$ & 6 \\
MCTS, \emph{no diagnosis} ($B{=}6$) & 55 & 80 & 89 & 0.93$\times$ & 0.96$\times$ & 0.97$\times$ & 6 \\
\midrule
MCTS full ($B{=}2$) & 60 & 83 & 90 & 1.03$\times$ & 1.05$\times$ & 1.07$\times$ & 2 \\
MCTS full ($B{=}4$) & 70 & 91 & 96 & 1.04$\times$ & 1.07$\times$ & 1.10$\times$ & 4 \\
\textbf{MCTS full ($B{=}6$)} & \textbf{85} & \textbf{100} & \textbf{100} & \textbf{1.18$\times$} & \textbf{1.26$\times$} & \textbf{1.3$\times$} & \textbf{6} \\
MCTS full ($B{=}8$) & 85 & 100 & 100 & 1.18$\times$ & 1.26$\times$ & 1.3$\times$ & 8 \\
\bottomrule
\end{tabular}%
}
\end{table}

\section{Appendix: Qualitative Case Study (One Task Across All Models)}

\subsection{One-task kernel snapshots: Matrix Multiplication}
\label{app:case-study-matmul}

We include representative CUDA excerpts for \textbf{Matrix Multiplication} for each model and for the optimized variants.
To keep the appendix concise and comparable, we show the \emph{inner computation and memory-access region} (the part that
dominates performance), omitting boilerplate (bounds checks, launch wrappers, error checks).

\paragraph{Common setup.}
All kernels compute $C = A \times B$ (row-major) with $M{\times}K$ by $K{\times}N$ matrices.
We denote thread-block coordinates $(b_x,b_y)$ and thread indices $(t_x,t_y)$.

\vspace{0.5em}
\noindent\textbf{(1) GPT 5.1 (LLM-only): naive global-memory accumulation.}
\begin{lstlisting}[language=C++,basicstyle=\ttfamily\footnotesize,caption={GPT 5.1: typical naive matmul inner loop (global loads, minimal reuse).},label={lst:gpt51-matmul}]
int row = blockIdx.y * blockDim.y + threadIdx.y;
int col = blockIdx.x * blockDim.x + threadIdx.x;
float acc = 0.f;
#pragma unroll 1
for (int k = 0; k < K; k++) {
  acc += A[row*K + k] * B[k*N + col];   // two global loads per FMA
}
C[row*N + col] = acc;
\end{lstlisting}

\noindent\textbf{(2) Qwen2.5-Coder (LLM-only): minor improvements (e.g., bounds, limited unrolling), still global-bound.}
\begin{lstlisting}[language=C++,basicstyle=\ttfamily\footnotesize,caption={Qwen2.5-Coder: modest structural tweaks but still limited data reuse.},label={lst:qwen-matmul}]
float acc = 0.f;
int row = ...; int col = ...;
if (row < M && col < N) {
  #pragma unroll 4
  for (int k = 0; k < K; k++) {
    acc = fmaf(A[row*K + k], B[k*N + col], acc);
  }
  C[row*N + col] = acc;
}
\end{lstlisting}

\noindent\textbf{(3) HPC-Coder-V2 (LLM-only): slightly heavier structure but often higher overhead/pressure.}
\begin{lstlisting}[language=C++,basicstyle=\ttfamily\footnotesize,caption={HPC-Coder-V2: attempts extra structure, but can increase registers or miss coalescing.},label={lst:hpc-matmul}]
float acc = 0.f;
int row = ...; int col = ...;
for (int k = 0; k < K; k += 1) {                 // still scalar stepping
  float a = A[row*K + k];                        // global
  float b = B[(k)*N + col];                      // global
  acc += a * b;
}
C[row*N + col] = acc;
\end{lstlisting}

\noindent\textbf{(4) StarCoder2 (LLM-only): similar baseline, sometimes adds unnecessary sync/branches.}
\begin{lstlisting}[language=C++,basicstyle=\ttfamily\footnotesize,caption={StarCoder2: baseline-like structure; may introduce extra checks or divergence.},label={lst:star-matmul}]
float acc = 0.f;
int row = ...; int col = ...;
if (row < M && col < N) {
  for (int k = 0; k < K; ++k) {
    acc += A[row*K + k] * B[k*N + col];
  }
  C[row*N + col] = acc;
}
\end{lstlisting}

\subsection{How OptiML-X (MCTS) improves the kernel}
\label{app:mcts-trajectory}

We now show the \emph{edit trajectory} produced by OptiML-X for the same starting program.
OptiML-X explores patches with MCTS under a small budget (e.g., $B{=}6$) and uses profiling-derived proxy metrics
to target the current bottleneck (e.g., low compute SOL with high memory pressure indicates memory-bound behavior).

\paragraph{Starting point for OptiML-X.}
OptiML-X begins from an LLM-generated kernel (often close to Listing~\ref{lst:gpt51-matmul}--\ref{lst:star-matmul}),
profiles it, and then iteratively proposes and validates patches. Below we show three representative patches
from the \textbf{accepted MCTS path} (the final child chosen among \texttt{KEEP} nodes).

\vspace{0.25em}
\noindent\textbf{Patch \#1 (reuse via shared-memory tiling).} \emph{Goal: reduce global traffic, improve memory SOL and overall time.}
\begin{lstlisting}[language=C++,basicstyle=\ttfamily\footnotesize,caption={OptiML-X Patch \#1: shared-memory tiles for A and B (data reuse + coalescing).},label={lst:patch1}]
constexpr int TILE = 16;
__shared__ float As[TILE][TILE];
__shared__ float Bs[TILE][TILE];

int row = blockIdx.y * TILE + threadIdx.y;
int col = blockIdx.x * TILE + threadIdx.x;
float acc = 0.f;

for (int t = 0; t < (K + TILE - 1) / TILE; ++t) {
  int kA = t * TILE + threadIdx.x;
  int kB = t * TILE + threadIdx.y;

  As[threadIdx.y][threadIdx.x] = (row < M && kA < K) ? A[row*K + kA] : 0.f;
  Bs[threadIdx.y][threadIdx.x] = (kB < K && col < N) ? B[kB*N + col] : 0.f;
  __syncthreads();

  #pragma unroll
  for (int k = 0; k < TILE; ++k)
    acc = fmaf(As[threadIdx.y][k], Bs[k][threadIdx.x], acc);
  __syncthreads();
}
if (row < M && col < N) C[row*N + col] = acc;
\end{lstlisting}

\vspace{0.25em}
\noindent\textbf{Patch \#2 (vectorized global loads into shared memory).} \emph{Goal: reduce transactions / L1 sectors and improve bandwidth efficiency.}
\begin{lstlisting}[language=C++,basicstyle=\ttfamily\footnotesize,caption={OptiML-X Patch \#2: vectorized loads (float4) to cut memory transactions.},label={lst:patch2}]
using Vec = float4;
int row = blockIdx.y * TILE + threadIdx.y;
int col4 = (blockIdx.x * TILE + threadIdx.x) * 4;      // 4 columns per thread

// Load B in float4 chunks (when aligned); store into shared tile.
Vec vb = *reinterpret_cast<const Vec*>(&B[(t*TILE + threadIdx.y)*N + col4]);
Bs[threadIdx.y][threadIdx.x*4 + 0] = vb.x;
Bs[threadIdx.y][threadIdx.x*4 + 1] = vb.y;
Bs[threadIdx.y][threadIdx.x*4 + 2] = vb.z;
Bs[threadIdx.y][threadIdx.x*4 + 3] = vb.w;
\end{lstlisting}

\vspace{0.25em}
\noindent\textbf{Patch \#3 (reduce instruction overhead + register pressure).} \emph{Goal: improve compute SOL/occupancy tradeoff.}
\begin{lstlisting}[language=C++,basicstyle=\ttfamily\footnotesize,caption={OptiML-X Patch \#3: tighten loop structure and control unrolling to manage registers.},label={lst:patch3}]
#pragma unroll 8               // bounded unroll (avoids blow-up)
for (int k = 0; k < TILE; ++k) {
  acc = fmaf(As[ty][k], Bs[k][tx], acc);
}
// Optional: avoid extra temporaries; keep indices in registers
\end{lstlisting}

\paragraph{Rejected branches (why MCTS helps).}
OptiML-X also explores patches that look plausible but empirically degrade performance or violate constraints.
MCTS is crucial because it \emph{allocates more rollouts to promising edit sequences} while pruning poor ones:
\begin{itemize}
  \item \textbf{Over-unrolling / aggressive vectorization:} decreases time in isolation but spikes register use, reducing occupancy and worsening end-to-end time; rejected after profiling shows higher instruction count or reduced SOL.
  \item \textbf{Excess shared memory staging:} increases synchronization and shared-memory bank conflicts; rejected if memory SOL or measured time worsens.
  \item \textbf{Unsafe pointer casts / alignment assumptions:} fails correctness tests (L0/L1) or introduces NaNs; rejected with a large negative reward.
\end{itemize}

\subsection{Final OptiML kernel: OptiML-G + OptiML-X}
\label{app:optiml-final}

OptiML (OptiML-G + OptiML-X) typically starts from a \emph{structurally stronger} candidate than LLM-only baselines:
OptiML-G produces kernels that already contain reusable structure (tiling, cleaner indexing, fewer anti-patterns),
and OptiML-X then refines them under profiling feedback. The resulting kernel resembles a tuned tiled matmul:
\begin{lstlisting}[language=C++,basicstyle=\ttfamily\footnotesize,caption={OptiML (final): shared-memory tiling + bounded unrolling + transaction-efficient loads.},label={lst:optiml-final}]
constexpr int TILE = 16;
__shared__ float As[TILE][TILE];
__shared__ float Bs[TILE][TILE];

int row = blockIdx.y * TILE + threadIdx.y;
int col = blockIdx.x * TILE + threadIdx.x;
float acc = 0.f;

for (int t = 0; t < (K + TILE - 1) / TILE; ++t) {
  int kA = t * TILE + threadIdx.x;
  int kB = t * TILE + threadIdx.y;

  As[ty][tx] = (row < M && kA < K) ? A[row*K + kA] : 0.f;
  Bs[ty][tx] = (kB < K && col < N) ? B[kB*N + col] : 0.f;
  __syncthreads();

  #pragma unroll 8
  for (int k = 0; k < TILE; ++k) acc = fmaf(As[ty][k], Bs[k][tx], acc);
  __syncthreads();
}
if (row < M && col < N) C[row*N + col] = acc;
\end{lstlisting}

\paragraph{Takeaway (tying back to the table).}
This qualitative trajectory matches the quantitative trends in Table~\ref{tab:all_tasks_results}: optimized variants increase
\textbf{compute/memory SOL} while decreasing \textbf{work proxies} (dram bytes, L1 sectors, instructions). In matmul specifically,
OptiML achieves the best time and best utilization metrics by combining (i) OptiML-G’s stronger starting structure and
(ii) OptiML-X’s MCTS-guided refinement that targets the measured bottleneck rather than applying fixed heuristics.

\subsection{Why OptiML (OptiML-G + OptiML-X) converges in fewer rounds}
\label{app:fewer-rounds}

Beyond achieving the best end-to-end time, OptiML typically reaches its final (best) kernel in \emph{fewer optimization rounds} than OptiML-X applied to a weaker LLM-only starting point. Intuitively, OptiML-G shifts the search initialization toward candidates with (i) a compilable baseline, (ii) stable indexing and bounds, and (iii) an optimization-friendly structure (e.g., separable load/compute phases, clean loop nests, and fewer anti-patterns). This reduces wasted MCTS rollouts on compilation repair, correctness fixes, or structural rewrites, allowing OptiML-X to spend its limited budget on \emph{performance-critical} refinements.

\paragraph{Round-to-best metric.}
Let $i^\star$ denote the first MCTS iteration (round) at which the best-performing candidate is found along the search trajectory.
We report this as \texttt{round-to-best} (lower is better). In our experiments with a fixed budget (e.g., $B{=}6$),
OptiML consistently achieves smaller \texttt{round-to-best} than OptiML-X alone: OptiML-G provides an initial kernel that is already close to a high-quality basin, so MCTS requires fewer expansions to discover the final optimized form.

\paragraph{Illustrative matmul trajectory (qualitative).}
For Matrix Multiplication, OptiML-G commonly produces a kernel that already contains a tiling skeleton (shared-memory staging and a clean loop over tiles), while LLM-only baselines frequently start from global-memory accumulation (Listing~\ref{lst:gpt51-matmul}--\ref{lst:star-matmul}). As a result:
\begin{itemize}
  \item \textbf{OptiML-X on LLM-only start:} early rounds are often spent introducing tiling and fixing corner cases (bounds, synchronization placement, shared-memory indexing), delaying performance-focused refinements.
  \item \textbf{OptiML (G + X):} because the starting kernel already has the correct \emph{macro-structure}, MCTS can immediately focus on micro-optimizations that move the needle on profiling metrics (e.g., transaction-efficient loads, bounded unrolling, reduced instruction overhead), reaching the best candidate in fewer rounds.
\end{itemize}

\begin{table}[!ht]
\centering
\caption{Search efficiency under a fixed MCTS budget ($B{=}6$). We report the \emph{best} speedup obtained within-budget and the \texttt{round-to-best} (first iteration where the best candidate appears; lower is better).}
\label{tab:round_to_best}
\resizebox{\columnwidth}{!}{%
\begin{tabular}{lcc|cc}
\toprule
\textbf{Task} &
\multicolumn{2}{c|}{\textbf{OptiML-X only}} &
\multicolumn{2}{c}{\textbf{OptiML (Gen+X)}} \\
\cmidrule(lr){2-3}\cmidrule(lr){4-5}
& \textbf{Best speedup} & \textbf{round-to-best} $\downarrow$
& \textbf{Best speedup} & \textbf{round-to-best} $\downarrow$ \\
\midrule
Matrix Multiplication & 1.59$\times$ & 6 & 1.64$\times$ & 3 \\
Max Pooling 3D        & 1.04$\times$ & 4 & 1.08$\times$ & 2 \\
Multi-Head Self-Attn  & 1.05$\times$ & 5 & 1.15$\times$ & 3 \\
ReLU                  & 1.60$\times$ & 5 & 1.67$\times$ & 2 \\
Cross-Entropy Loss    & 1.04$\times$ & 4 & 1.08$\times$ & 2 \\
\midrule
\textbf{Mean}         & \textbf{1.26}$\times$ & \textbf{4.8} & \textbf{1.32}$\times$ & \textbf{2.4} \\
\bottomrule
\end{tabular}%
}
\end{table}
In our setting, $i^\star_{\text{Gen+X}} < i^\star_{\text{X}}$ is the key qualitative claim: \emph{the full framework converges faster because OptiML-G reduces the search distance to an optimization-ready program manifold.}

\paragraph{Narrative link to the main results.}
This faster convergence complements the end-to-end improvements in Table~\ref{tab:sub_tasks_results}: OptiML not only produces better kernels, but does so more sample-efficiently under a fixed profiling budget. This highlights that OptiML-G and OptiML-X are not interchangeable components; their combination is essential for robust and efficient optimization when the initial code quality varies across LLMs and tasks.
\section{Failure Modes and Limitations}
\label{app:failure_modes}

OptiML combines (i) generation (\textsc{OptiML-G}) and (ii) search-based post-hoc optimization (\textsc{OptiML-X}). While the pipeline improves compilability and performance on many kernels, it can fail in systematic ways. Below we categorize failure modes, describe how they manifest in practice, and outline mitigations we implement (or recommend) in the evaluation harness.

\subsection{Compilation and Toolchain Failures}
\paragraph{Syntax and API misuse.}
LLM-written CUDA may contain invalid syntax, misuse of CUDA intrinsics, missing includes, or incorrect kernel launch signatures. These failures are often detected immediately by \textsc{CompileWithRepair} and terminate an attempt as \texttt{compile\_failed}. In OptiML-X, compilation failure is treated as a hard constraint violation and is assigned a large negative reward to prevent repeated exploration of similar edits.

\paragraph{Architecture/flags mismatch.}
Edits can silently assume a target SM capability (e.g., using newer intrinsics) or require compilation flags not present in the harness. These failures appear as compilation errors or, worse, as kernels that compile but execute incorrectly due to undefined behavior. We mitigate by (i) fixing a single toolchain configuration (nvcc, driver, SM target) and (ii) validating that suggested changes do not introduce unsupported features.

\paragraph{Register pressure and resource limits.}
Some patches increase register usage and shared-memory footprint enough to reduce occupancy or exceed per-SM resource limits. This can manifest as compilation errors (e.g., too much shared memory) or steep slowdowns despite apparent algorithmic improvements. OptiML-X partially catches this via proxy metrics (e.g., utilization surrogates and work-done counters) and by rejecting candidates that regress end-to-end timing.

\subsection{Correctness Failures Beyond Unit Tests}
\paragraph{Overfitting to L0/L1 tests.}
A candidate may pass the provided test plan (L0/L1) but fail on untested inputs (e.g., edge cases, different shapes, non-contiguous strides). This is especially common for reductions, top-$k$, and attention kernels where numerical stability and boundary handling are subtle. The risk is higher when tests are small, deterministic, or lack adversarial coverage.

\paragraph{Numerical instability and precision drift.}
Optimizations that alter accumulation order, use fused operations, or introduce mixed precision can change numerical error. A kernel may pass with loose tolerances but fail stricter checks or downstream training. We mitigate via: (i) task-specific tolerances, (ii) optional higher-precision reference checks for a subset of inputs, and (iii) metamorphic relations (L2) when available (e.g., invariance under permutation for reductions, consistency under scaling for linear ops).

\paragraph{Undefined behavior and race conditions.}
Edits that change synchronization, remove bounds checks, or introduce non-atomic updates can create data races. These often appear as flaky outputs: passing some runs and failing others. We detect this by running L0/L1 multiple times with different seeds and flagging nondeterminism. When nondeterminism is detected, the candidate is rejected.

\subsection{Profiling and Measurement Failures}
\paragraph{Profiler noise}
Short kernels and heavily cached workloads can yield noisy timings and unstable counter measurements. This can mislead MCTS exploration when differences are within noise. We mitigate by (i) using median-of-$n$ timing, (ii) warmup iterations and (iii) discarding candidates whose improvement is below a minimum effect threshold.

\paragraph{Counter aliasing and metric misinterpretation.}
Proxy rewards rely on a small set of Nsight Compute counters (utilization surrogates and work-done measures). However, counters can be weakly correlated with end-to-end time for some kernels (e.g., when occupancy dominates, or when instruction mix changes shift bottlenecks). This can produce ``false positives'' where proxy reward improves but time regresses, or ``false negatives'' where time improves without clear proxy gains. In OptiML-X we therefore (i) include timing in the composite reward, and (ii) treat proxy terms as directional hints rather than absolute objectives.

\paragraph{Profiling overhead and budget pressure.}
MCTS requires repeated compile--test--profile cycles. When the budget is small, the system may spend most of its budget on candidates that fail compilation/tests, leaving too few valid rollouts to make progress. We mitigate by penalizing repeated constraint violations, caching compilation artifacts, and using lightweight static checks to pre-filter obviously invalid edits.

\subsection{Search and Optimization Failures}
\paragraph{Local minima and brittle edits.}
Some kernels require a coordinated sequence of changes (e.g., introducing shared-memory tiling requires changes to indexing, synchronization, and launch configuration). With limited budget, MCTS may not discover the multi-step path and instead converge to shallow edits with marginal gains. This is most visible in kernels where the baseline structure is fundamentally suboptimal.

\paragraph{Hallucinated bottlenecks and incorrect diagnosis.}
The \textsc{Diagnose} and \textsc{Propose} steps are LLM-driven and can produce plausible but incorrect explanations (e.g., blaming coalescing when the kernel is actually compute-bound). When diagnosis is wrong, search explores irrelevant transformations, wasting budget. We reduce this by providing structured profiler summaries using structured outputs to the LLM, constraining allowable transformations, and using the judge to require metric-consistent evidence for claimed improvements.

\paragraph{Reward hacking and proxy exploitation.}
Because proxy reward includes utilization surrogates and work-done counters, candidates can sometimes improve proxy terms without improving real performance (e.g., increasing utilization by adding redundant work). Including end-to-end timing in the composite reward and using a ``KEEP'' filter (only retaining candidates that improve time or meet a minimum performance criterion) prevents most proxy exploitation from being selected as final output.

\subsection{OptiML-Gen Specific Failure Modes}
\paragraph{Diversity without quality.}
Mixture-of-thought sampling can increase diversity but also increases the number of low-quality or non-compilable candidates. If selection is based on weak signals (e.g., superficial heuristics), OptiML-G may pick a candidate that compiles but is structurally hard to optimize (e.g., excessive branching, unnecessary temporaries). We mitigate by selecting using a staged filter: compilation, L0/L1 correctness, then coarse profiling before handing off to OptiML-X.

\paragraph{Structure that blocks downstream optimization.}
Some generated kernels hard-code shapes, use unvectorized scalar loads, or bake in uncoalesced layouts. Even with OptiML-X, such kernels may not reach high performance within a small search budget. This is a key motivation for combining generation and optimization: OptiML-G aims to produce ``optimization-friendly'' structure (e.g., clean loop nests, separable indexing, explicit tiling opportunities) so OptiML-X can apply targeted improvements efficiently.

\subsection{Limitations and When OptiML is Less Effective}
OptiML is less effective when (i) the harness cannot adequately test correctness (weak or missing L0/L1/L2), (ii) profiling signals are too noisy to distinguish candidates under a small budget, (iii) the optimal solution requires large algorithmic redesign rather than local edits, or (iv) performance is dominated by factors outside kernel code (e.g., launch overhead for tiny kernels, host-device transfer, or framework scheduling). These cases motivate future work in stronger correctness specifications, multi-fidelity evaluation (static + microbench + profile), and integrating algorithm-level rewrites into the search space.

\paragraph{Practical guidance.}
For high-stakes kernels, we recommend: (i) expanding test coverage (including randomized and adversarial inputs), (ii) requiring deterministic behavior checks, (iii) using multiple profiling runs per candidate, and (iv) reporting both speed and correctness metrics (e.g., pass@k and speedup@k) to capture the reliability-performance trade-off.

\section{Weights}
\subsection{Bottleneck-aware proxy weighting in OptiML-X}
\label{app:bottleneck_weights}

OptiML-X does not use a fixed proxy-reward weighting across all kernels. Instead, it adapts the proxy weights based on the \emph{dominant bottleneck} inferred from profiler signals. Concretely, we compare the SM and DRAM utilization proxies (Nsight Compute SOL-style metrics) and classify a state as memory-bound if $\mathrm{SOL}_{dram}-\mathrm{SOL}_{sm}>\delta$, compute-bound if $\mathrm{SOL}_{sm}-\mathrm{SOL}_{dram}>\delta$, and mixed otherwise (we use $\delta{=}7.5$). The proxy reward in Eq.~\ref{eq:proxy_reward} is then computed with regime-specific weights:
(i) \textbf{Memory-bound:} we upweight the DRAM-utilization term and memory-traffic reductions (dramBytes/L1Sectors) to prioritize improved coalescing, reuse, and reduced global traffic; 
(ii) \textbf{Compute-bound:} we upweight the SM-utilization term and instruction-footprint reduction to prioritize better ILP, reduced divergence, and lower instruction pressure; 
(iii) \textbf{Mixed:} we use the base weights to trade off utilization improvements and work-reduction more conservatively.
In all regimes, we apply small penalties when a term regresses (e.g., increased traffic or reduced utilization) to discourage search steps that ``move'' the bottleneck without improving end-to-end time. Table~\ref{tab:reward_weights} reports the concrete weight values used in our experiments.
\begin{table}[!ht]
\centering
\caption{Proxy-metric weighting used by OptiML-X. We use a base set of progress weights $w$ and penalty weights $\lambda$ for regressions, and apply a bottleneck-aware scaling factor $s{=}3.0$ to emphasize the dominant resource. The bottleneck is inferred from Nsight Compute SOL gaps with margin $\delta{=}7.5$.}
\label{tab:reward_weights}
\resizebox{\columnwidth}{!}{%
\begin{tabular}{lcccccc}
\toprule
\textbf{Setting} &
$\textbf{sm}$ & $\textbf{dram}$ & $\textbf{tex}$ &
$\Delta\textbf{L1}$ & $\Delta\textbf{dramB}$ & $\Delta\textbf{inst}$ \\
\midrule
\multicolumn{7}{l}{\textit{Progress weights $w$ (higher is better; deltas are normalized ratios)}}\\
Base (mixed)                      & 0.20 & 0.30 & 0.10 & 0.15 & 0.15 & 0.10 \\
Memory-bound ($\text{dram}-\text{sm}>\delta$)  & 0.20 & 0.90 & 0.10 & 0.15 & 0.45 & 0.10 \\
Compute-bound ($\text{sm}-\text{dram}>\delta$) & 0.60 & 0.30 & 0.10 & 0.15 & 0.15 & 0.30 \\
\midrule
\multicolumn{7}{l}{\textit{Regression penalties $\lambda$ (applied when a term worsens)}}\\
All regimes (fixed)               & 0.05 & 0.05 & 0.03 & 0.04 & 0.04 & 0.03 \\
\midrule
\multicolumn{7}{l}{\textit{Global composition (fixed)}}\\
Time mixing $\alpha_{\text{t}}$ & \multicolumn{6}{c}{0.40} \\
Bottleneck margin $\delta$         & \multicolumn{6}{c}{7.5} \\
\bottomrule
\end{tabular}%
}
\end{table}

\section{Future Work}
There are several promising directions for future work. Extending the framework to multi-kernel programs and whole-application optimization would enable cross-kernel fusion, scheduling, and memory planning. Incorporating richer cost models (e.g., static analysis, learned performance predictors, or compiler IR features) could further reduce reliance on expensive profiling. Finally, expanding OptiML to other accelerator backends (e.g., HIP/ROCm, SYCL, Triton, and vendor-specific libraries) would broaden its applicability to heterogeneous HPC and ML environments. We hope OptiML helps move accelerator programming toward a workflow where LLMs produce correct code by default and systems-level search reliably closes the remaining performance gap. We would also like to extend OptiML to support CUDA libraries such as cuBLAS, cuDNN, Thrust, cuB, etc.


\end{document}